\documentclass[journal]{IEEEtran}
\usepackage{graphics} 
\usepackage{epsfig} 
\usepackage{times} 
\usepackage{amsmath} 
\usepackage{amssymb}  
\usepackage{cite}
\usepackage{epstopdf}
\usepackage{subfigure,color,balance}
\usepackage{verbatim}
\usepackage{cases}
\usepackage{enumerate}
\usepackage{booktabs}
\usepackage{balance}
\usepackage{mathbbol}
\usepackage{algorithm}
\usepackage{algorithmicx}
\usepackage{algpseudocode}
\usepackage{comment}
\usepackage{gensymb}
\usepackage{textcomp}
\usepackage{threeparttable}
\usepackage{multirow}
\usepackage{makecell}

\usepackage{tikz}
\newcommand*{\circled}[1]{\lower.7ex\hbox{\tikz\draw (0pt, 0pt)%
		circle (.5em) node {\makebox[1em][c]{\small #1}};}}
\usepackage[colorlinks=true,      
linkcolor=black,      
citecolor=black,      
filecolor=black,      
urlcolor=blue]{hyperref}
\usepackage{array}

\newtheorem{remark}{Remark}

\newcolumntype{C}{>{\centering\arraybackslash}p{0.55cm}}

\def\0{{\bf 0}}
\def\1{{\bf 1}}

\begin{document}
%
\title{
{
Multi-Source Human-in-the-Loop Digital Twin Testbed  for Connected and Autonomous Vehicles \\in Mixed Traffic Flow
}

}

\author{Jianghong Dong, Chunying Yang, Mengchi Cai, Chaoyi Chen,
\\Qing Xu, Jianqiang Wang, Jiawei Wang, and Keqiang Li 
	\thanks{Jianghong Dong, Chunying Yang, Mengchi Cai, Chaoyi Chen, Qing Xu, Jianqiang Wang and Keqiang Li are with the School of Vehicle and Mobility, Tsinghua University, Beijing 100084, China. (djh20@mails.tsinghua.edu.cn,  ycyacademic@gmail.com, \{caimengchi,chency2023, qingxu,wjqlws,likq\}@tsinghua.edu.cn) }
     \thanks{Jiawei Wang was with the School of Vehicle and Mobility, Tsinghua University, Beijing 100084, China. He is now with the Department of Civil and Environmental Engineering, University of Michigan, Ann Arbor, MI 48109, USA. (jiawe@umich.edu)}

 %
	%
}

\maketitle

\begin{abstract}
In the emerging mixed traffic environments, Connected and Autonomous Vehicles (CAVs) have to interact with surrounding human-driven vehicles (HDVs). This paper introduces MSH-MCCT (Multi-Source Human-in-the-Loop Mixed Cloud Control Testbed), a novel CAV testbed that captures complex interactions between various CAVs and HDVs. Utilizing the Mixed Digital Twin concept, which combines Mixed Reality with Digital Twin, MSH-MCCT integrates physical, virtual, and mixed platforms, along with multi-source control inputs. Bridged by the mixed platform, MSH-MCCT allows human drivers and CAV algorithms to operate both physical and virtual vehicles within multiple fields of view. Particularly, this testbed facilitates the coexistence and real-time interaction of physical and virtual CAVs \& HDVs, significantly enhancing the experimental flexibility and scalability. Experiments on vehicle platooning in mixed traffic showcase the potential of MSH-MCCT to conduct CAV testing with multi-source real human drivers in the loop through driving simulators of diverse fidelity. The videos for the experiments are available at our project website: \url{https://dongjh20.github.io/MSH-MCCT}. 
\end{abstract}


\section{Introduction}

The rapidly advancing and evolving technology of Connected and Autonomous Vehicles (CAVs) promises to enhance travel safety, traffic efficiency, and environmental friendliness of transportation systems~\cite{ding2025patterns,dong2025stfc,shi2024collaborative}. To validate CAVs' potential before their practical deployment, they must undergo rigorous and comprehensive testing and evaluation~\cite{feng2023dense,feng2021intelligent,zhou2025hierarchical}. Particularly, since achieving full automation and networking will take a considerable amount of time, CAVs have to interact with diverse human-driven vehicles (HDVs) in a foreseeable future~\cite{wang2020controllability,zheng2020smoothing}. Therefore,  the upcoming traffic system is essentially a human-in-the-loop mixed-autonomy system~\cite{gao2025discussion}, which makes it crucial to conduct human-in-the-loop testing and evaluation for CAVs in mixed traffic environments, where multiple multi-source HDVs coexist~\cite{wang2023implementation,yue2022effects,wang2021digital}.

To conduct such CAV testing in mixed traffic, one straightforward approach is real-world experiments, where HDVs could be naturally involved. Despite its high potential in obtaining authentic driving behavior of human drivers~\cite{maag2012studying}, this approach generally takes substantial labor and material costs, and is extremely time- and space-inefficient~\cite{liao2021cooperative}. Moreover, experiments involving safety-critical scenarios are unsuitable to be carried out in the real world due to safety concerns. Alternatively, to incorporate human driver factors, existing studies mostly focus on simulation-based testing by introducing human driver models to capture HDVs' behaviors~\cite{chen2021mixed,wu2021flow,wang2022make}, such as intelligent driver model (IDM)~\cite{kesting2010enhanced} for car-following maneuvers, Gipps’s lane changing decision model~\cite{gipps1986model} for lane-changing maneuvers, and social interaction models~\cite{wang2025multiagent} based on imitation learning.
Recently, researchers have begun to explore the use of artificial neural networks to simulate driver models~\cite{hao2024empowering,liu2023can}. 
Although these models are typically derived from extensive datasets of human driver behaviors~\cite{ma2024driver}, but their accuracy and efficiency for CAV testing remain challenged by the inherently uncertain and nonlinear characteristics of human behaviors. 

\subsection{Human-in-the-Loop Experiments and Digital Twin}

Human-in-the-loop experiments via driving simulators bridge the gap between field experiments and numerical simulations. 
Typical simulator hardware ranges from high-fidelity commercial products~\cite{warner2017simulator} to simplified configurations with computer screens, steering wheels and pedals~\cite{wang2021digital}. 
The driving environment is commonly provided by the software of driving simulators~\cite{lindner2022coupled}, including commercial program such as SILAB~\cite{maag2012studying} and open-source ones such as Unity~\cite{wang2021digital2} and CARLA~\cite{Dosovitskiy2017}. 
In most cases, the driving environment is a virtual one, which is highly configurable and reproducible~\cite{abdelgawad2017networked}. 
Since this approach is affordable and easily implementable, and its results is empirically proved to be valid~\cite{swake2013driver}, it has 
already been widely adopted for data collection and behavior modeling of human drivers~\cite{wang2023towards,hu2022review,abdelgawad2017networked,negash2023driver}.

Recently, this approach has also been utilized for human-in-the-loop testing and validation for CAV technologies in mixed traffic environments~\cite{wang2021digital,dong2023mixed}. 
Precisely, human drivers operate virtual vehicles in a virtual environment via driving simulators, while the CAVs are controlled by specific CAV algorithms and interact with the HDVs in the same virtual environment. 
Most existing research considers a simplified setup, where only one single HDV is controlled by a human driver via a driving simulator, and the other traffic participants are programmed~\cite{wang2021digital,warner2017simulator,wang2021digital2}. 
Very recently, to better approximate and investigate the complex multi-interactivity nature associated with the advanced automotive technologies and future mixed traffic environments~\cite{abdelgawad2017networked}, an experimental platform is constructed in~\cite{yue2022effects} with multiple HDVs in the loop, where the effects of CAV's merging maneuvers on surrounding HDVs' behaviors are evaluated.
Specifically, multiple driving simulators are connected together, and thus drivers can interact simultaneously in the same virtual driving environment, where the CAV is controlled by algorithms. 
Indeed, this setup is typical in multi-driver driving simulation~\cite{maag2012studying}, also known as networked driving simulation~\cite{abdelgawad2017networked}, which aims to explore the interactions between multiple road users in real-world traffic. 

\begin{figure}[t]
	\vspace{1mm}
	\centering
	\subfigure[Classical DT]
	{\includegraphics[scale=0.55]{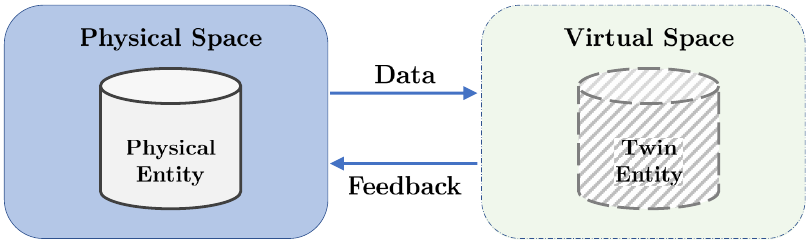}
		\label{fig.dt-architecture}}
	\hspace{4mm}
	\subfigure[MixedDT]
	{\includegraphics[scale=0.55]{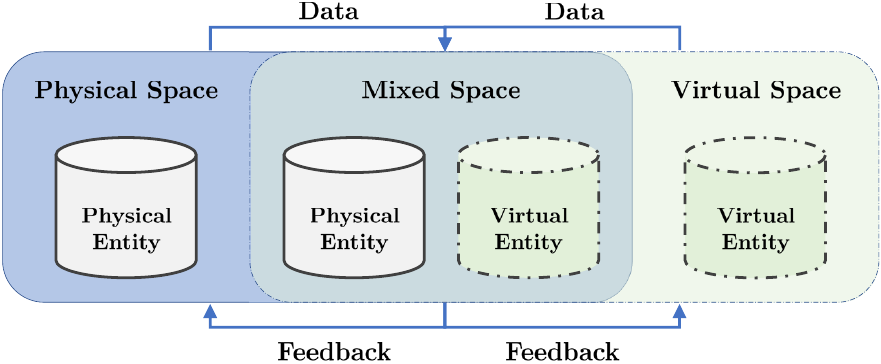}
		\label{fig.mdt-architecture}}
	\caption{Schematics for classical DT and mixedDT. (a) In classical DT, the virtual space is typically a digital replica of the physical space. (b) In mixedDT, the virtual space and the physical space are integrated into the mixed space, where physical and virtual entities could coexist and interact with each other.
	}
	\label{fig.architecture}
  \vspace{-4mm}
\end{figure}

However, the method in~\cite{yue2022effects,maag2012studying,abdelgawad2017networked} is constrained by the inherent limitation of the underlying virtual simulation environment, where many real-world factors cannot be accurately replicated, such as vehicle dynamics, uncertainties in sensor measurement, and delays in communication and computation processes. 
To address this limitation and integrate physical factors into the original virtual environments, 
Digital Twin (DT) emerges as a promising tool, which facilitates the connection between physical and virtual entities~\cite{glaessgen2012digital,zhu2019parallel,fei2024critical,bae2025digital}. 

As shown in Fig.~\ref{fig.dt-architecture}, DT comprises two main parts: physical entities in the physical space, and the virtual counterparts of the physical entities (named as twin entities) in the virtual space~\cite{grieves2014digital}. 
Through synchronization and interaction between real-world entities and their virtual counterparts, DT promises to facilitate CAV testing via this bio-connection~\cite{liao2021cooperative,wang2022mobility,ma2024driver}. 
Furthermore, DT serves as a critical enabling technology for the emerging metaverse in transportation~\cite{zhou2025vetaverse} and parallel driving~\cite{wang2017parallel}.
Recently, some CAV testing experiments with human in the loop  have been conducted in typical applications based on DT, including adaptive cruise control~\cite{wang2021digital2}, cooperative ramp merging~\cite{liao2021cooperative} and intersection control~\cite{wang2021digital}. However, only one single HDV controlled by a real human driver is integrated into the DT environment in these research, which cannot reflect the complex multi-interactivity nature between multiple CAVs and multiple HDVs in future mixed traffic environments. 
Moreover, classical DT approaches typically regard virtuality as a digital replica of reality, and consequently, the potential of virtual entities in CAV testing remains underexplored in previous studies~\cite{wang2021digital2,liao2021cooperative,wang2021digital}. 

Therefore, the fundamental challenge persists in integrating experimental physical and virtual vehicles—as well as control commands from both human drivers and algorithms, each operating independently in isolated environments—into a unified digital twin environment where they can coexist and interact in real-time to enable effective multi-driver-in-the-loop CAV testing.
To address these issues, this paper aims to establish a more flexible and scalable platform for CAV testing with multi-source real human drivers in the loop based on a very recent notion, called Mixed Digital Twin (mixedDT)~\cite{dong2023mixed}.

\subsection{Notion of Mixed Digital Twin (mixedDT)}
As an extension of the classical DT concept~\cite{glaessgen2012digital,grieves2017digital}, mixedDT consists of three spaces: physical space, virtual space, and mixed space, as illustrated in Fig.~\ref{fig.mdt-architecture}. The meanings of the physical space and the virtual space are consistent with the classical DT, while the newly added mixed space is an integration of the physical and virtual spaces, where physical and virtual entities coexist and interact in real time. 
Particularly, the virtual entity in mixedDT exists independently in the virtual space, rather than a digital replica or counterpart of the physical entity, \emph{i.e.}, twin entity in the classical DT. Therefore, virtual entities could coexist and interact with physical entities in the mixed space, contributing to enhanced flexibility and scalability. Based on this new notion, our previous work establishes a preliminary platform, called Mixed Cloud Control Testbed (MCCT)~\cite{dong2023mixed}. 
MCCT bridges multiple diverse entities from the vehicle-road-cloud integrated system~\cite{chu2021cloud} into a comprehensive experimental platform via mixedDT. 
Along this direction, this work extends this platform to support multi-source human-in-the-loop experiments, and thus the complex interactions between multiple CAVs and HDVs in upcoming mixed traffic environments can be captured and investigated. 

\begin{table*}[t!]
\centering
\caption{Comparison of CAV Testing Platforms Highlighting Multi-Human-in-the-Loop Capabilities
}
\label{tab-platform_comparison_grid}
\setlength{\tabcolsep}{4pt}
\renewcommand{\arraystretch}{1.3} 
\resizebox{\textwidth}{!}{
\begin{tabular}{| l | CCC | CCC | CCC | CCC | CCC | CCC |}
\hline
\multirow{3}{*}{Metrics} & \multicolumn{18}{c|}{Platforms} \\
\cline{2-19}

& \multicolumn{3}{c|}{SUMO} 
& \multicolumn{3}{c|}{CARLA} 
& \multicolumn{3}{c|}{\makecell{NVIDIA\\Drive Sim}} 
& \multicolumn{3}{c|}{VeHIL} 
& \multicolumn{3}{c|}{Mcity} 
& \multicolumn{3}{c|}{\textbf{MSH-MCCT}} \\
\cline{2-19}

& L & M & H 
& L & M & H 
& L & M & H 
& L & M & H 
& L & M & H 
& L & M & H \\
\hline

Support for Multiple Human Drivers 
& $\bullet$ & & 
& & $\bullet$ & 
& & $\bullet$ & 
& $\bullet$ & & 
& & & $\bullet$
& & & $\bullet$ \\
\hline

Physical-Virtual Integration 
& $\bullet$ & & 
& $\bullet$ & & 
&  & $\bullet$ & 
& & & $\bullet$
& & & $\bullet$ 
& & & $\bullet$ \\
\hline

Scalability 
& & & $\bullet$ 
& & $\bullet$ &  
& & $\bullet$ &   
& $\bullet$ & & 
& $\bullet$ & & 
& & $\bullet$ & \\
\hline

Affordability 
& & & $\bullet$ 
& & & $\bullet$ 
&  & $\bullet$ & 
& $\bullet$ & & 
& $\bullet$ & & 
& & $\bullet$ & \\
\hline

Reproducibility 
& & & $\bullet$ 
& & & $\bullet$ 
& & & $\bullet$  
& & $\bullet$ & 
& $\bullet$ & & 
& & $\bullet$ & \\
\hline

\multicolumn{19}{l}{\footnotesize *Note: L = Low, M = Medium, H = High. This comparison focuses solely on the native capabilities of the platforms, excluding any further custom development.} \\
\end{tabular}
}
\end{table*}

A related concept with mixedDT is Mixed Reality (MR), which is defined as a class of simulators combining both physical and virtual objects to create a hybrid of the physical and virtual spaces~\cite{rokhsaritalemi2020review}. MR evolves from multiple concepts including Augmented Reality and Augmented Virtuality. It aims to construct the “Reality-Virtuality Continuum” and consequently provides consumers a more immersive and interactive environment~\cite{flavian2019impact}.
Recently, MR has been applied to CAV testing and evaluation to extend the existence of interactive objects in not only virtual testing environments~\cite{serrano2023digital} but also physical testing environments~\cite{feng2020safety}. To address the practical testing demands of CAVs, mixedDT introduces the idea of MR into the classical DT by establishing a mixed space where both physical and virtual entities could coexist and interact with each other. Note that the physical space is a real-world testing environment, while the virtual space is highly configurable and reproducible, which is particularly applicable in safety-critical situations. The mixed space bridges the physical and virtual spaces to operate together, and thus provides more potential applications, enhancing the flexibility and scalability of DT's applications in CAV testing.

\begin{figure}[t]
	\centering
	\includegraphics[scale=0.9]{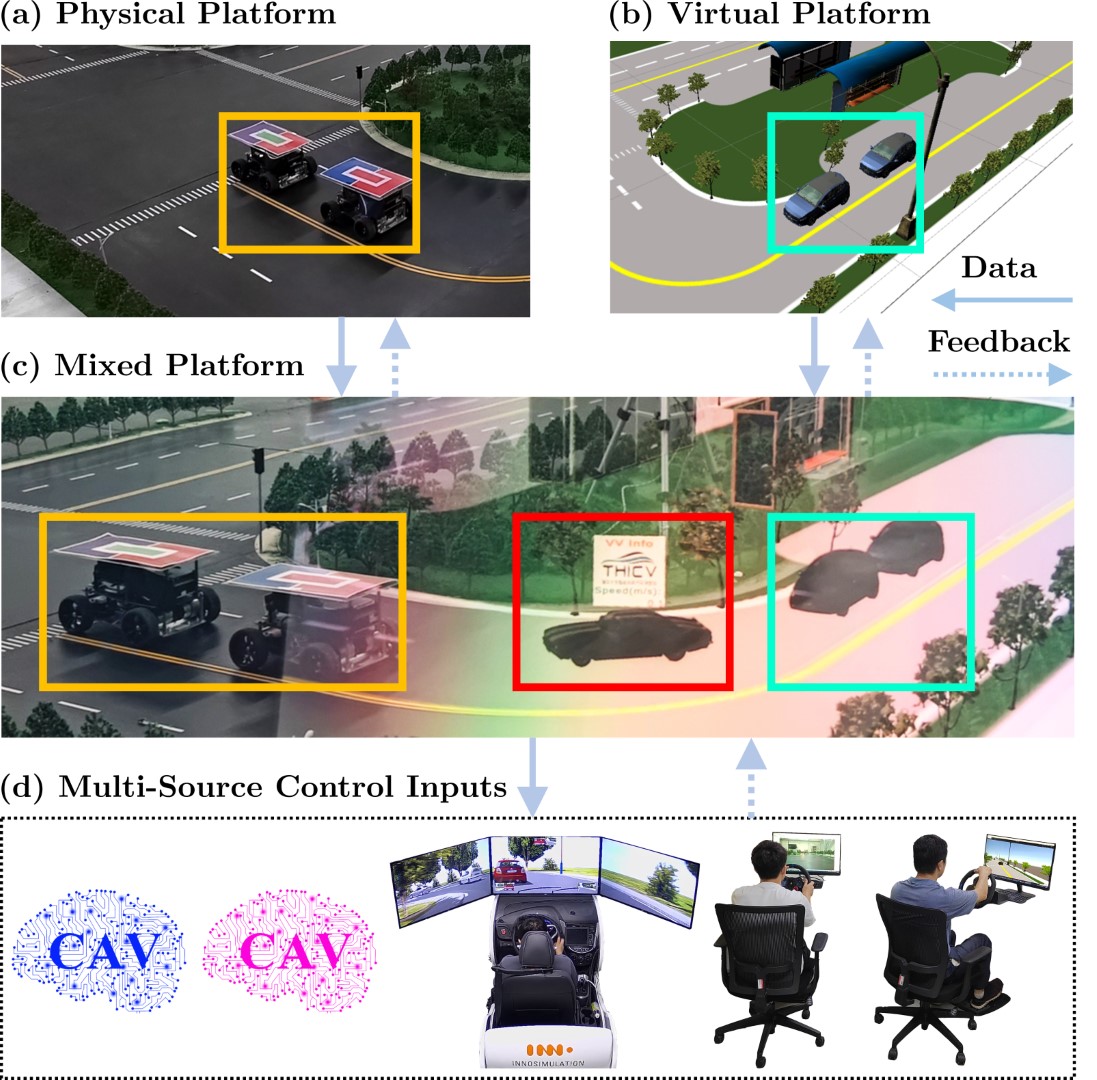}
	\caption{Schematic of MSH-MCCT. In the mixed platform, the physical vehicles in the physical platform (outlined in gold), the virtual vehicles in the virtual platform (outlined in cyan) and the virtual vehicle in the driving environment of a driving simulator (outlined in red) coexist and interact with each other in real-time. 
    In contrast, in existing platforms~\cite{zayas2021digital,zayas2021ids}, physical and virtual vehicles operate exclusively within their respective environments, without effective integration or interactive mechanisms established between them.
    The multi-source control inputs includes human drivers via various-fidelity levels of driving simulators and diverse CAV algorithms. The visualization of the mixed platform shown in (c) is provided by a MR device. The demonstration videos are available at our project website: \url{https://dongjh20.github.io/MSH-MCCT}.} 
	\label{fig.MSH-framework}
	\vspace{-4mm}
\end{figure}

\subsection{Contributions}

Building on the foundational MCCT platform~\cite{dong2023mixed}, this paper introduces the advancements made with MSH-MCCT (Multi-Source Human-in-the-Loop Mixed Cloud Control Testbed), a novel extended platform for CAV testing.
Employing the mixedDT framework, MSH-MCCT serves for experimental testing and validation of CAV technologies with multi-source real human drivers in the loop. As shown in Fig.~\ref{fig.MSH-framework}, MSH-MCCT consists of four core components: physical platform, virtual platform, mixed platform, and multi-source control inputs. The physical platform provides a miniature real-world testing environment, while the virtual platform brings flexibility for the quantity, motion behavior and dynamics of involved vehicles. The multi-source control inputs include human drivers via various-fidelity levels of driving simulators, and diverse CAV algorithms.
In particular, physical CAVs \& HDVs and virtual CAVs \& HDVs coexist and interact simultaneously within a unified and integrated environment, \emph{i.e.}, the mixed platform, which is deployed on cloud servers, utilizing cloud computing technologies to enhance computational efficiency and storage capabilities for users~\cite{armbrust2010view}.

\begin{figure*}[t]
	\centering
	\includegraphics[width=0.99\textwidth]{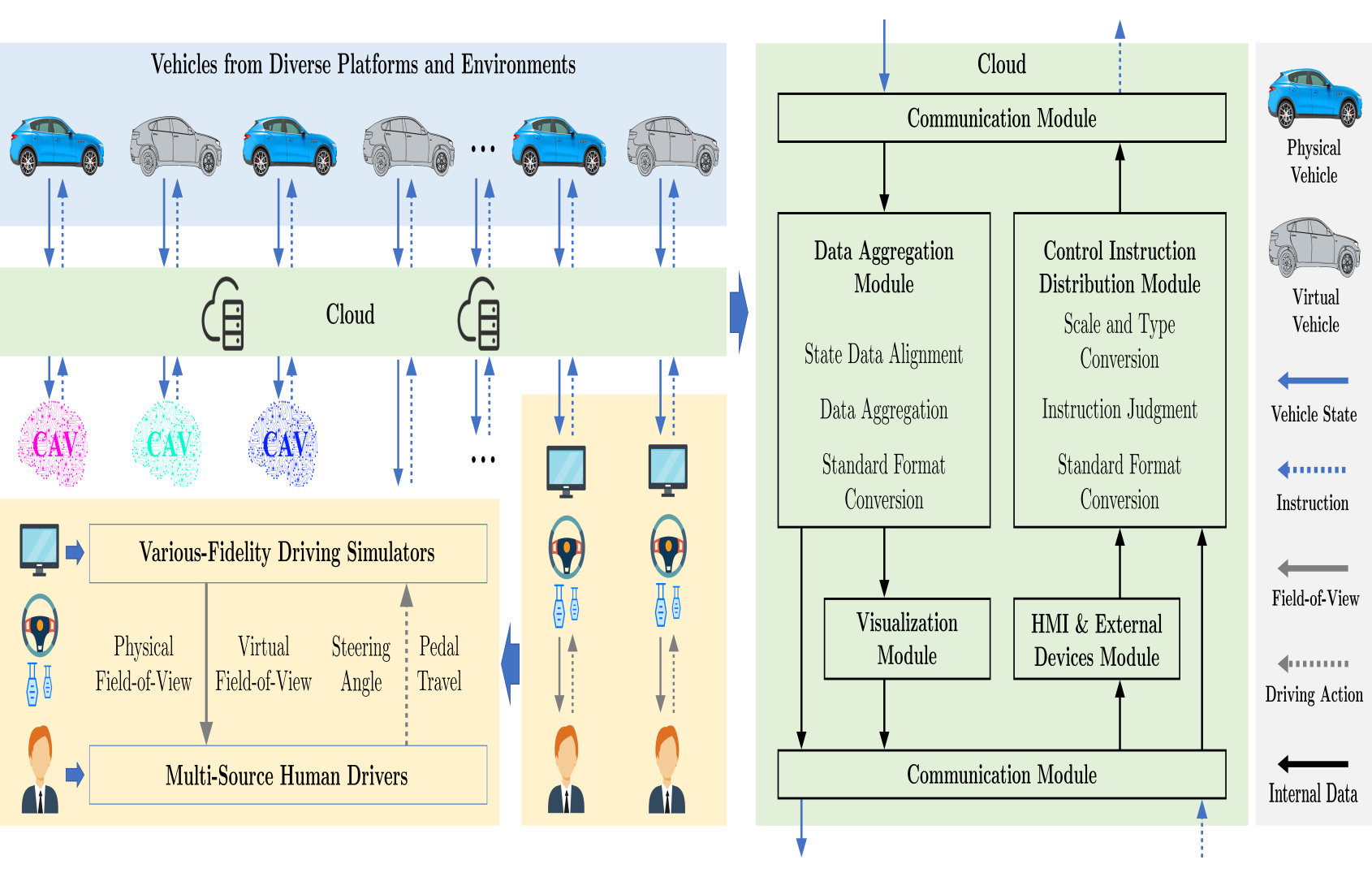}
	\vspace{-2mm}
	\caption{The methodological framework for conducting CAV testing with multi-source human drivers in the loop via driving simulators in MSH-MCCT. Both physical and virtual vehicles could be controlled by human drivers and CAV algorithms through a cloud-based intermediary. Consequently, the four types of vehicles, \emph{i.e.}, physical HDVs, virtual HDVs, physical CAVs and virtual CAVs, coexist and interact with each other simultaneously within the mixed platform.}
	\label{fig.MSH-method}
	\vspace{-4mm}
\end{figure*}

Conversely, in existing platforms~\cite{zayas2021digital,zayas2021ids}, physical and virtual vehicles operate exclusively within their respective environments, without effective integration or interactive mechanisms established between them. Table.~\ref{tab-platform_comparison_grid} compares MSH-MCCT with several representative CAV testing platforms in terms of their support for multi-human-driver-in-the-loop experiments. 
		 The evaluated platforms can be categorized into three types: (1) purely virtual environments, including SUMO~\cite{lopez2018microscopic} for macroscopic traffic flow simulation, as well as CARLA~\cite{Dosovitskiy2017} and NVIDIA Drive Sim~\cite{li2024choose} for microscopic high-fidelity rendering; (2) virtual-physical integrated environments, represented by VeHIL~\cite{gietelink2006development} and our MSH-MCCT; and (3) a purely physical environment, i.e., Mcity~\cite{xu2021system}.
Furthermore, the comparison dimensions consist of the following: support for multi-human-driver-in-the-loop experiments (Support for Multiple Human Drivers); the capability of physical-virtual integration (Physical-Virtual Integration); scalability in terms of vehicle quantity, behavior, and dynamics (Scalability); the affordability of platform setup and operation (Affordability); and the reproducibility of experimental results (Reproducibility).
The comparative results reveal the superiority of our MSH-MCCT platform in facilitating multi-human-driver-in-the-loop experiments.
Our main contributions and features of MSH-MCCT are specified as follows.

\begin{itemize}
	\item To our best knowledge, MSH-MCCT is the first DT testing platform for CAVs which integrates multiple real human drivers simultaneously into the loop. Unlike existing platforms, which primarily rely on simulated driver behaviors~\cite{wu2021flow,yan2023learning,wang2023deep} or limited real human interactions~\cite{wang2021digital,warner2017simulator,wang2021digital2}, MSH-MCCT supports real-time interactions among an arbitrary number of CAVs and HDVs operated by real human drivers, facilitated by the mixedDT framework~\cite{dong2023mixed}. This allows for a realistic representation of complex CAVs-HDVs interactions in emergent mixed traffic scenarios. 
	
	\item MSH-MCCT enables human drivers and CAV algorithms to control both physical and virtual vehicles via a cloud-based intermediary. In addition, drivers are provided with dual perspectives: the physical field-of-view from on-board sensors and the virtual field-of-view from the twin vehicle. This feature offers a comprehensive experience that is unattainable in traditional single-type (\emph{e.g.}, physical vehicles~\cite{la2012development,weiss2022high} or virtual vehicles~\cite{wang2021digital,maag2012studying,wang2021digital2}) or single-perspective platforms (\emph{e.g.},  physical view~\cite{kannapiran2020go} or virtual view~\cite{oeltze2015benefits,wang2020augmented}). Our approach addresses typical challenges associated with ``organizational siloing"~\cite{grieves2017digital} and ``cyber-physical fusion"~\cite{tao2018digital} in DT applications.

	\item We evaluate the platform's effectiveness with vehicle platooning experiments in mixed traffic flow, where cooperative adaptive cruise control (CACC)~\cite{milanes2013cooperative} is selected as the testing algorithm for CAVs and multi-source real human drivers are involved. The experimental results not only demonstrate the algorithm’s ability to mitigate traffic perturbations, but also capture collision events in safety-critical scenarios, which is non-trivial to achieve in real-world tests and often poorly replicated in simulations. This validates the MSH-MCCT's robust capabilities for realistic and scalable CAV testing in mixed traffic with multiple real human drivers in the loop. 
\end{itemize}

The rest of this paper is organized as follows. Section~\ref{sec.2} presents the overall architecture of MSH-MCCT. Section~\ref{sec.3} introduces detailed methodology based on mixedDT, while Section~\ref{sec.4} elaborates the implementation details. Experimental setup and results are presented in Section~\ref{sec.5}, and Section~\ref{sec.6} concludes this paper.

\section{Platform Architecture}
\label{sec.2}
In this section, we introduce the overall architecture and methodology framework of MSH-MCCT. As shown in Fig.~\ref{fig.MSH-framework}, MSH-MCCT consists of four parts: 
\begin{enumerate}
 \setlength\itemsep{0.1em}
    \item Physical platform, which is constructed on a sand table and provides a miniature real-world testing environment. The physical vehicles run on the physical platform.
    \item Virtual platform, which is developed by a game engine. The virtual vehicles operate on the virtual platform, whose quantity, motion behavior and dynamics can be flexibly designed. 
    \item Multi-source control inputs, including human drivers and CAV algorithms. 
The driving actions performed by human drivers based on the driving field-of-view in driving simulators are measured by the steering wheels and pedals. The CAV algorithms deployed on the cloud or external hosts return control instructions based on the vehicle state information.
\item Mixed platform, which is deployed in cloud servers. It bridges the physical platform, the virtual platform and the multi-source control inputs into an integrated whole, where they achieve synchronous operation and cross-platform real-time interaction. 
\end{enumerate}

The methodological framework of MSH-MCCT is shown in Fig.~\ref{fig.MSH-method}. Multiple CAVs and HDVs, each of which could be both physical and virtual, coexist and interact with each other in an integrated environment, \emph{i.e.}, the mixed platform. 
As the central component of MSH-MCCT, the cloud transmits vehicle state information from diverse platforms and environments to both the human drivers and the CAV algorithms, and subsequently relays the control instructions from them to the corresponding vehicles for execution. Precisely, the cloud aligns and aggregates vehicle state data sent by both physical and virtual vehicles, and then transmits this processed state data to the driving simulators and the hosts where the CAV algorithms are deployed. 
The driving simulators render the driving field-of-view based on received data and collect corresponding control inputs by the human drivers. The CAV algorithms also generate control instructions based on the received data.  The cloud distributes received control instructions to the corresponding vehicles for execution after necessary processing. 
\begin{figure*}[t]
	\centering
	\subfigure[The physical field-of-view]
	{\includegraphics[scale=0.22 ]{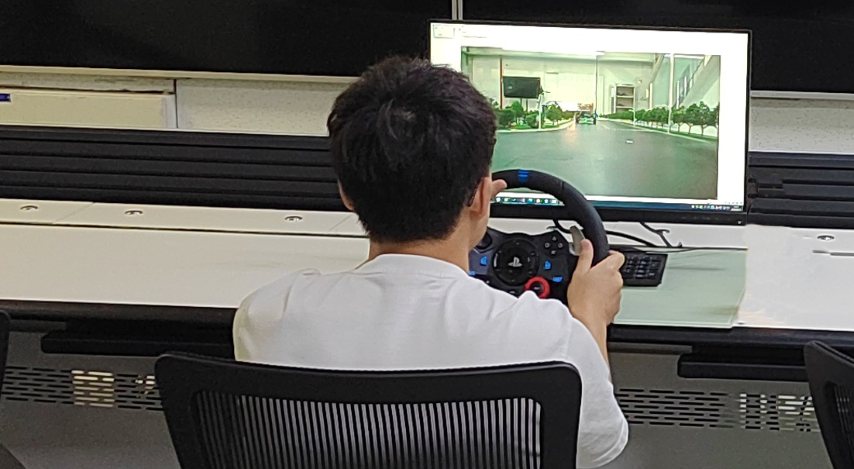}
	\label{fig.physical view}
	}
	\subfigure[The virtual field-of-view]
	{\includegraphics[scale=0.22 ]{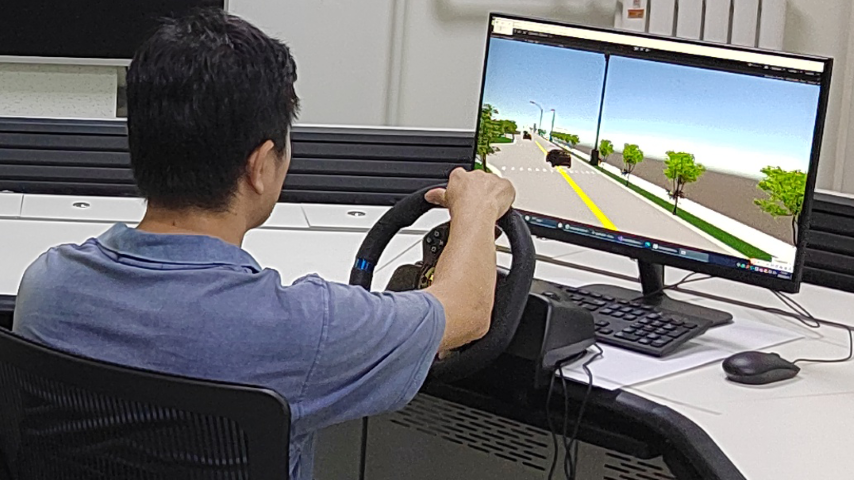}
	\label{fig.virtual view}
	}
    \subfigure[The speed prompt panel]
	{\includegraphics[scale=0.22 ]{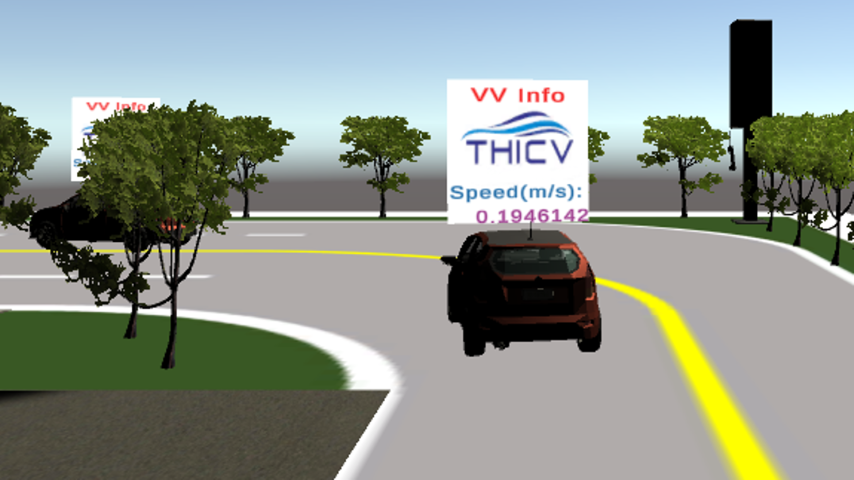}
	\label{fig.speed promt}
	}
	\vspace{-1mm}
	\caption{Driving field-of-view and prompt panel. In (a) and (b), the snapshots of human-in-the-loop experiments based on physical and virtual field-of-view are presented respectively; see the screens for the specific view. In (c), the real-time speed prompt of the  preceding vehicle in the virtual driving field-of-view is displayed on the white panel on the top of the preceding vehicle.}
	\label{fig.view and promot}
  \vspace{-4mm}
\end{figure*}

Basically, the cloud bridges the vehicles from diverse platforms and environments and the multi-source control inputs to achieve synchronous operation and cross-platform real-time interaction. 
From the aforementioned operation workflow, the cloud also serves as the core and backbone of MSH-MCCT. As illustrated in the right part of Fig.~\ref{fig.MSH-method}, the cloud mainly consists of five functional modules: 1)
The communication module is responsible for information exchange between the cloud and other entities in the platform, as well as external devices. 2)
The data aggregation module aligns and aggregates all the vehicle state data from diverse platforms and environments, and provides standardized formatted vehicle state data to be used by other modules and entities. 3) 
The control instruction distribution module processes and distributes all the received control instructions from multi-source human drivers and diverse CAV algorithms, which are then transmitted to the corresponding vehicles by the communication module. 
4) The visualization module utilizes the outcomes from the data aggregation module to accomplish the visualization of the mixed platform in both physical and virtual spaces.
5) The Human-Machine Interface (HMI) and external devices module refers to interacting with entities in MSH-MCCT via HMI devices such as a MR head-mounted display and external devices such as personal computers. The methodology and the practical implementations will be covered in the Sections~\ref{sec.3} and~\ref{sec.4}.

\section{Methodology}
\label{sec.3}
In this section, we introduce the methodology for conducting CAV testing with multi-source human drivers in the loop via driving simulators in MSH-MCCT. Particularly, we elaborate on how the HDVs are controlled by human drivers via various-fidelity driving simulators. From the driver's perspective, we explain in detail how to drive physical and virtual vehicles in the platform. In addition, we present how CAVs and HDVs are integrated to operate together in the mixed platform. 

\subsection{Driving a Physical Vehicle}

Since many practical factors cannot be accurately replicated in the simulations~\cite{la2012development,kannapiran2020go}, such as vehicle dynamics, uncertainties in sensor measurement, and delays in communication and computation processes, we design the mode of driving physical vehicles via driving simulators based on mixedDT. 
The physical vehicles run on the physical driving environment, which is a $1:14$ miniature sand table in MSH-MCCT with respect to the real world, and each vehicle uploads its state data, such as velocity and acceleration, to the cloud and then executes received driver's instructions from the cloud to update its own status. 

Based on the driving field-of-view delivered by screens, the driver operates the driving simulator, including the steering wheel, the accelerator and brake pedals, which can be precisely measured via the supporting software or Software Development Kit (SDK) of the driving simulator. 
The simulator communicates with the cloud via wired connection to ensure real-time performance, and exchanges the driver’s control input signals and necessary data. 
Particularly,  we develop two kinds of driving field-of-view: physical field-of-view and virtual field-of-view.

\subsubsection{Driving based on physical field-of-view}

As shown in Fig.~\ref{fig.physical view}, the physical field-of-view observed by drivers is the real-time image captured by the on-board camera in the physical vehicle, which depicts the real-world scene in the physical platform. The image is uploaded to the cloud by the physical vehicle, and then received by the driving simulator to be subsequently displayed on the screen and visible to the driver. 
Since the driving simulator is commonly fixed-base and the physical vehicle is moving, this physical field-of-view via on-board cameras is typically employed for driving physical vehicles through simulators~\cite{la2012development,kannapiran2020go}. 

However, since it is non-trivial to combine the virtual environment (particularly virtual vehicles) with the view from the on-board camera, this physical field-of-view is not capable to show the existence and movement of virtual vehicles. 
In addition, when the physical vehicle is a miniature one, which is common due to its flexibility and safety in recent research for CAV testing~\cite{kannapiran2020go,chalaki2022research}, obtaining authentic driving reactions from the driver becomes more challenging due to the difference between miniature driving environment and real-world driving environment. 

\subsubsection{Driving based on virtual field-of-view}
In this kind, the virtual driving field-of-view from the twin vehicle of the corresponding physical vehicle is utilized to be displayed to the driver; see Fig.~\ref{fig.virtual view} for visualization. 
Since the status of the twin vehicle and the corresponding physical vehicle are synchronized in real time, it is only necessary to display the virtual driving field-of-view of the twin vehicle in the virtual environment to the driver, which is easily accessible in simulation software such as Unity and CARLA. 
Precisely, the driving simulator continuously receives state information of all the vehicles from the cloud, and subsequently re-renders the driving field-of-view on screens. 
Within the virtual field-of-view, all the physical vehicles are represented by their corresponding virtual twin vehicles. Hence, both physical and virtual vehicles can be simultaneously displayed to the driver, and thus supports experiments in the mixed space where both vehicles are involved. This resolves the aforementioned problems encountered by the physical field-of-view. 

Moreover, the virtual field-of-view is highly flexible and scalable, allowing the incorporation of various virtual driving prompts and guidance~\cite{liu2021vision}. 
Our MSH-MCCT platform also leverages this feature. One example is illustrated in Fig.~\ref{fig.speed promt}, where the real-time speed of the preceding vehicle is displayed to the human driver.

\begin{remark}[A novel paradigm for human-in-the-loop driving simulations] 
``Cyber-physical fusion'' is regarded as one of the most critical challenges for DT applications~\cite{grieves2017digital}. 
Very recently, a Virtual Reality (VR) device has been utilized in~\cite{weiss2022high,xu2022analyzing} to provide the driver with a virtual driving field-of-view, while a full-size physical vehicle is employed to provide the driver with real-world physical feedback.
For human-in-the-loop driving simulation via driving simulator, existing work mostly utilizes the simulator to control either a physical vehicle based on a physical field-of-view~\cite{la2012development,kannapiran2020go} or a virtual vehicle based on a virtual field-of-view~\cite{wang2021digital,wang2020augmented}.
Our mixedDT-inspired approach simultaneously utilizes the real physical vehicle dynamics and the flexible and scalable virtual driving field-of-view by incorporating the virtual twin vehicle of the corresponding physical vehicle.  Consequently, the fusion of the physical vehicle and the virtual environment is achieved: the physical vehicle executes the driver's control instructions in the physical environment, and meanwhile the perceptual information of the physical vehicle in the mixed environment is provided virtually to the driver. This presents a novel paradigm for conducting human-in-the-loop driving simulations via driving simulators.
\end{remark}

\subsection{Driving a Virtual Vehicle}

In this mode, a driver operates a virtual vehicle in a virtual driving environment via a driving simulator. 
While miniature physical vehicles can introduce real-world uncertainties and disturbances, their dynamics still differ from those of full-size vehicles. Therefore, it is necessary to incorporate virtual vehicles with flexibly configurable dynamics.
Moreover, compared with physical driving environments, virtual environments enable easier and more cost-effective reproduction of both common and safety-critical scenarios~\cite{maag2012studying}.

In MSH-MCCT, the virtual vehicles and the virtual driving environment are developed based on a game engine, and they communicate with the cloud through wired connection to ensure real-time performance. The virtual driving environment, aligning with the real-world environment in terms of spatial extent, is a scaled digital replica of the physical driving environment (miniature physical sand table). 
This forms the basis for quickly and accurately aligning the positional coordinates of the physical and virtual vehicles within the mixed platform, which are from the physical and virtual platforms respectively, elaborated in Sections~\ref{sec.integration in mixed} and~\ref{sec.mixed platform}.
Similar to the physical vehicle, the virtual vehicle sends its state data to the cloud and executes received driver's control instructions. 
Furthermore, the first-person driving field-of-view of any virtual vehicle as illustrated in Fig.~\ref{fig.virtual view} is easily accessible in the game engine. 
The remaining steps to drive a virtual vehicle in the virtual driving environment are the same as driving a physical vehicle based on virtual field-of-view.

\begin{figure}[t]
	\centering
	\includegraphics[scale=0.8]{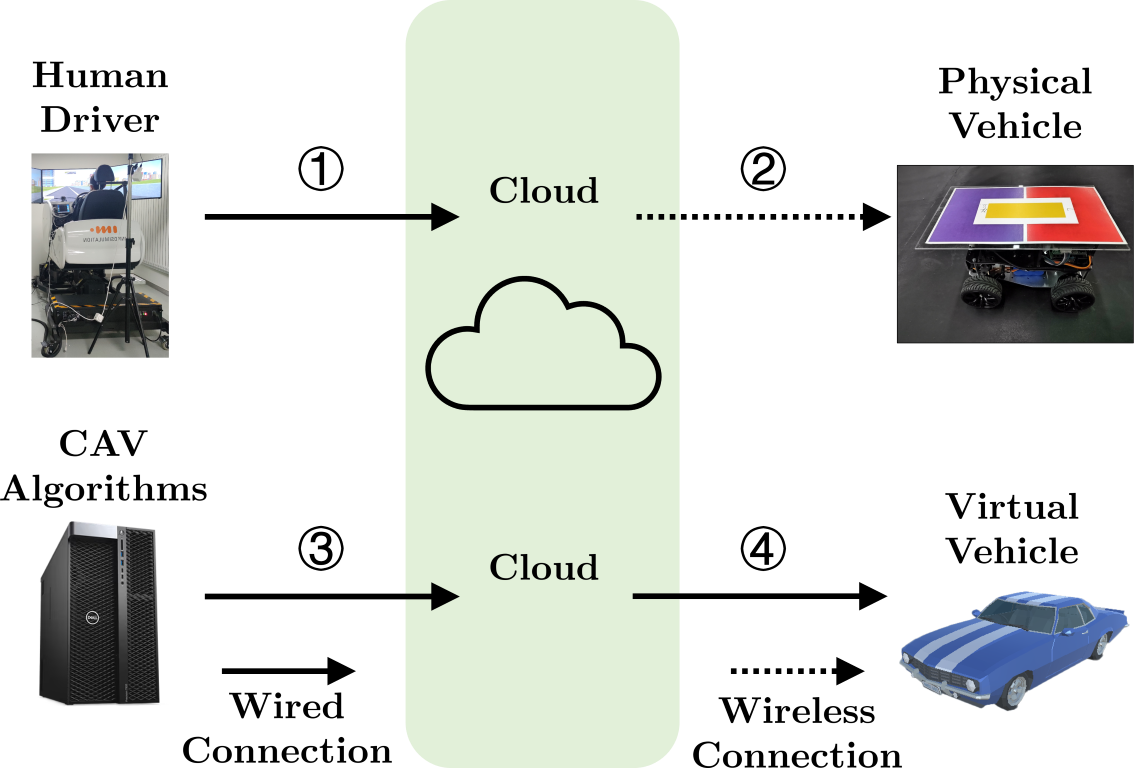}
	\vspace{-2mm}
	\caption{Control modes diagram. Human drivers and CAV algorithms could control both the physical vehicles and the virtual vehicles via a cloud-based intermediary.}
	\label{fig.control mode}
	\vspace{-2mm}
\end{figure}

\begin{figure*}[t]
	\centering
	\subfigure[Physical driving environment]
	{\includegraphics[scale=0.8]{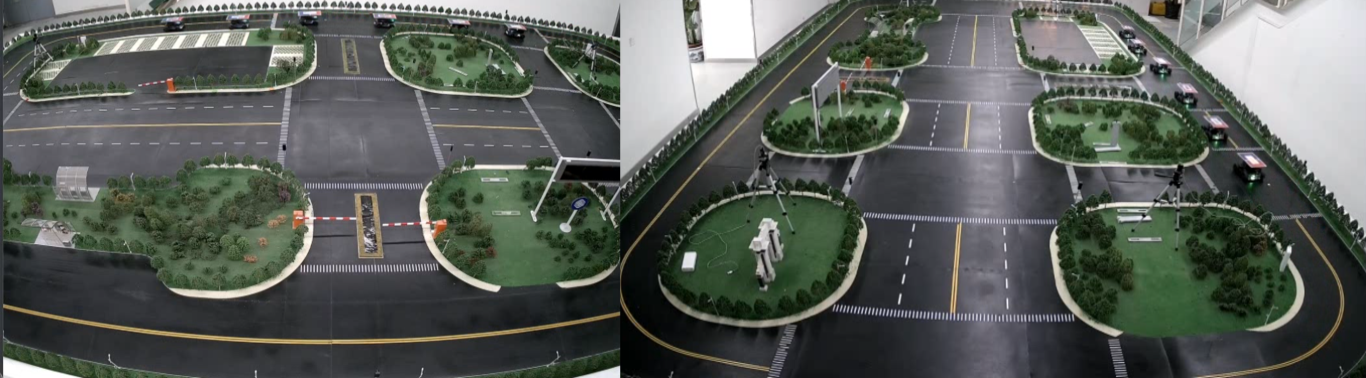}
		\label{fig.physical environment}}
    \subfigure[Virtual driving environment]
	{\includegraphics[scale=0.5]{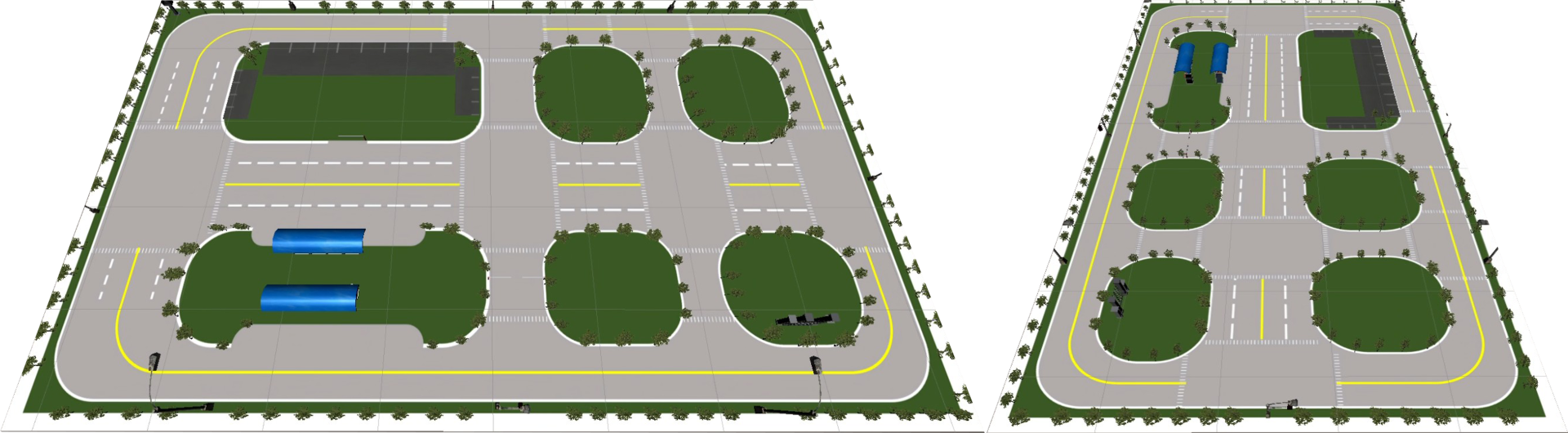}
		\label{fig.virtual environment}}
	\subfigure[Physical vehicle]
	{\includegraphics[scale=0.8]{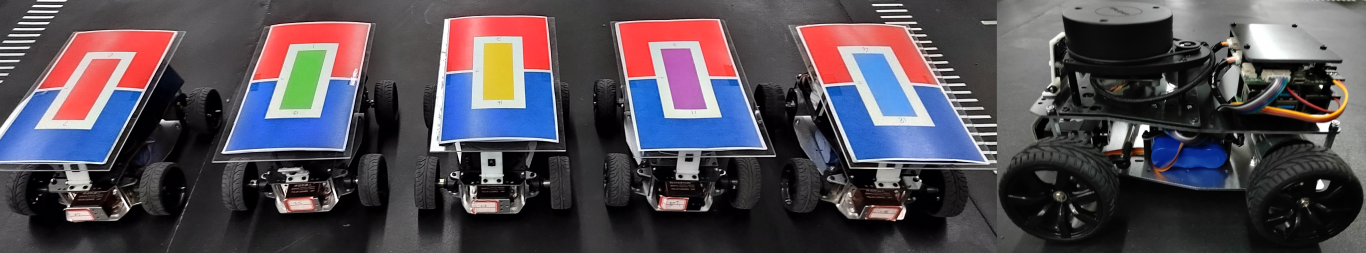}
		\label{fig.physical vehicle}}
    \subfigure[Virtual vehicle]
	{\includegraphics[scale=0.65]{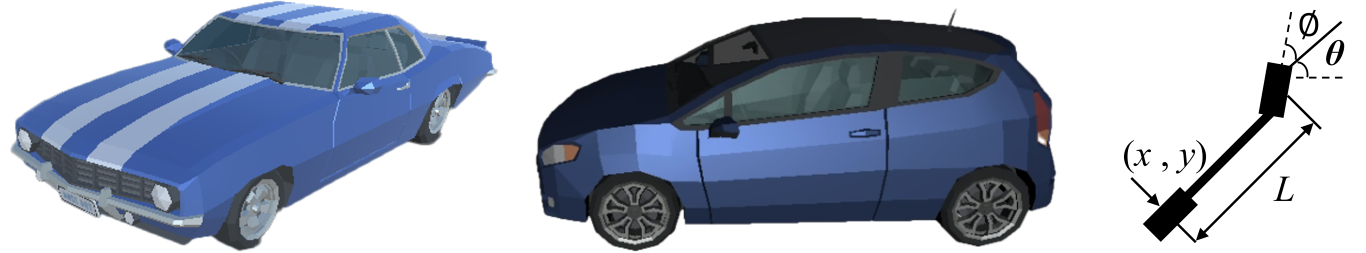}
		\label{fig.virtual vehicle}}
	\vspace{-2mm}
	\caption{Physical and virtual driving environments and vehicles. (a) The physical sand table, \emph{i.e.}, the physical driving environment, having a size of $9\,\mathrm{m} \times 5\,\mathrm{m}$. (b) The virtual sand table, \emph{i.e.}, the virtual driving environment, developed based on the Unity game engine. (c) The miniature physical vehicle. A color block card is pasted to the top for localization. (d) Two commonly used vehicle body models and the default dynamics model of the virtual vehicles, \emph{i.e.}, the typical bicycle model~\cite{rajamani2011vehicle}. 
	}
	\label{fig.physical platform}
  \vspace{-4mm}
\end{figure*}

\begin{table*}[t]
\footnotesize
\caption{Platform Setups}
\label{tab.platform parameters}
\centering
		\begin{threeparttable}
		\setlength{\tabcolsep}{10mm}{
			\begin{tabular}{cc}
   \toprule
				Component & Setup \\\hline
				Physical sand table & $9\,\mathrm{m} \times 5\,\mathrm{m}$,  $1:14$ with respect to the real-world driving environment \\
                Roadside sensors in the physical platform & Four identical cameras in a $2\times 2$ manner with overlap \\
                Roadside facilities in the physical platform & $73$ street lamps, $11$ traffic lights and $3$ lifting rods \\
                Miniature physical  vehicles & $215\,\mathrm{mm} \times 190\,\mathrm{mm} \times 125\,\mathrm{mm}$, powered by Raspberry Pi 4B\\
                Host of the virtual platform & Processor: Intel Core i7-10700K CPU @ 3.80 GHz $\times$ 16, Memory: 16 GB \\
                Host of the G29 simulator & Processor: Intel Core i7-10700 CPU@ 2.90 GHz $\times$ 16, Memory: 16 GB  \\
                Host of the InnoSimulation simulator & Processor: Intel Core i7-8700 CPU @ 3.20 GHz $\times$ 12, Memory: 16 GB  \\
                Cloud server for the mixed platform & Processor: 2 $\times$ Intel Xeon Gold 5220R CPU @ 2.20GHz $\times$ 48, Memory: 64 GB \\
                
				\bottomrule
		\end{tabular}}
  
		\end{threeparttable}
	\vspace{-2mm}
\end{table*}

\subsection{Integration of CAVs and HDVs in the Mixed Platform}
\label{sec.integration in mixed}

Diverse CAVs and HDVs are integrated in the mixed platform operating on the cloud.
As illustrated in Fig.~\ref{fig.control mode}, human drivers and CAV algorithms could control both physical and virtual vehicles through a cloud-based intermediary. 
In other words, the four types of vehicles, \emph{i.e.}, physical HDVs, virtual HDVs, physical CAVs and virtual CAVs, coexist and interact with each other simultaneously within the mixed platform. 
Precisely, the cloud integrates physical and virtual vehicles by the data aggregation module, and meanwhile, the cloud integrates human drivers and CAV algorithms via the control instruction distribution module, distributing their control instructions  to the vehicles for execution; see the right part of Fig.~\ref{fig.MSH-method} for illustration. 
The cloud bridges the physical vehicles and the virtual vehicles with the human drivers and the CAV algorithms to operate together within a unified, shared environment, \emph{i.e.}, the mixed platform. This achieves simultaneous operation and cross-platform real-time interaction between them, allowing for comprehensive exploration of complex interactions between multiple CAVs and HDVs associated with the future mixed traffic environment~\cite{abdelgawad2017networked}.

In the following, we present the details of the procedures in the cloud shown in the right part of Fig.~\ref{fig.MSH-method}. In the data aggregation module, the cloud initially aligns the state data such as the positional coordinates of all the vehicles from diverse platforms and environments, as the sizes and coordinate systems of different physical and virtual environments differ. 
Recall that the physical environment in MSH-MCCT is a $1:14$ miniature sand table with respect to the real world, while the virtual driving environment keeps consistent with the real world in terms of spatial extent. 
After corrections for communication delays, the state data of all the vehicles is aggregated, which is then transformed into a standardized format, usable for other modules such as the visualization module and authorized external devices. 
Consequently, the physical and virtual vehicles from the physical and virtual platforms respectively are integrated into a unified environment, and thus they can coexist and interact with each other. 
Essentially, this process twins the physical vehicles into the virtual space, enabling them to interact with the virtual vehicles indirectly via their respective digital counterparts.

In the control instruction distribution module, the cloud first performs scale and type conversions on all received control instructions from human drivers and CAV algorithms. 
Scale conversion is necessary due to differences of dimensions between different driving environments. 
Multi-source control inputs need to be converted into a unified control instruction type, \emph{i.e.}, desired front wheel angle and desired speed. 
Then, the cloud processes the control instructions by filtering out the instructions beyond a specific range in case of sudden excessive control inputs. 
Finally, the processed instructions are converted into a standardized format for transmission. After allocation by a vehicle-instruction correspondence mechanism, the communication module sends the instructions to the corresponding vehicles for execution. The aforementioned procedures are repeated in the cloud, and accordingly, human drivers and CAV algorithms are integrated into a unified environment. 


\begin{remark}[``Hot-swapping'' capability in MSH-MCCT] \label{Remark:Hotswapping}
In most existing driving simulation platforms, a driving simulator is typically associated with controlling one pre-assigned vehicle, which is non-trivial to change during the experiments~\cite{wang2021digital,kannapiran2020go}. 
Based on the mixedDT framework, MSH-MCCT achieves decoupling between controlled vehicles and driving simulators. 
Their correspondence can be dynamically configured as needed within the cloud at any time. After the alteration, the cloud forwards the driver's control instructions to the new corresponding vehicle for execution. Note that the driving field-of-view of each vehicle is available in the local simulation software. 
Consequently, drivers can switch between driving different vehicles smoothly during experiments, achieving a ``hot-swapping'' property in  MSH-MCCT. 
This significantly enhances the flexibility and scalability for conducting human-in-the-loop experiments via driving simulators. For instance, one single driving simulator can accomplish multiple types of driving simulations. 
Moreover, remote driving interventions can also be achieved for non-functional CAVs. 

\end{remark}

\section{Implementation Details of MSH-MCCT}
\label{sec.4}

In this section, we present the practical implementation details of the four components of MSH-MCCT, \emph{i.e.}, physical platform, virtual platform, multi-source control inputs and mixed platform. Also, we introduce the design for visualization, communication and interaction. Some fundamental setups of the platform are illustrated in Table~\ref{tab.platform parameters}.

\subsection{Physical Platform}

The physical platform represents the physical space in mixedDT, and is constructed on a sand table. It mainly consists of physical driving environment, physical vehicles and roadside sensors and facilities. 
The physical sand table shown in Fig.~\ref{fig.physical environment} provides the physical driving environment of MSH-MCCT, which is a controllable and affordable miniature real-world testing environment. 
The sand table contains two-way four-lane roads in the middle and two-way two-lane roads in both sides, as well as intelligent traffic lights. This basic road structure of the physical platform could support common CAVs  technologies, such as platoon control~\cite{wang2021leading}, multi-lane formation~\cite{cai2022formation},  intersection cooperation~\cite{hu2023lane} and traffic network management~\cite{lu2021optimization}. 

The physical vehicles in MSH-MCCT are the miniature vehicles shown in Fig.~\ref{fig.physical vehicle} and there are currently nine identical vehicles. The vehicle is rear-wheel drive and front-wheel steering with an Ackerman steering structure. The color block card on the top of each vehicle is recognized by the roadside cameras for localization.
The control paradigm of the physical vehicle adopts a cloud-based networked control architecture~\cite{xu2023cloud}. 
Precisely, the physical vehicle subscribes to the control instructions published in the Robot Operating System (ROS) topic message form by the cloud, and then executes it to achieve desired velocity and steering angle. Under this architecture, the execution and calculation of the control instructions are fully decoupled, which supports the aforementioned ``hot-swapping'' property in Remark~\ref{Remark:Hotswapping}. 

\begin{figure}[t]
	\vspace{1mm}
	\centering
	\subfigure[G29 simulator]
	{\includegraphics[scale=0.3]{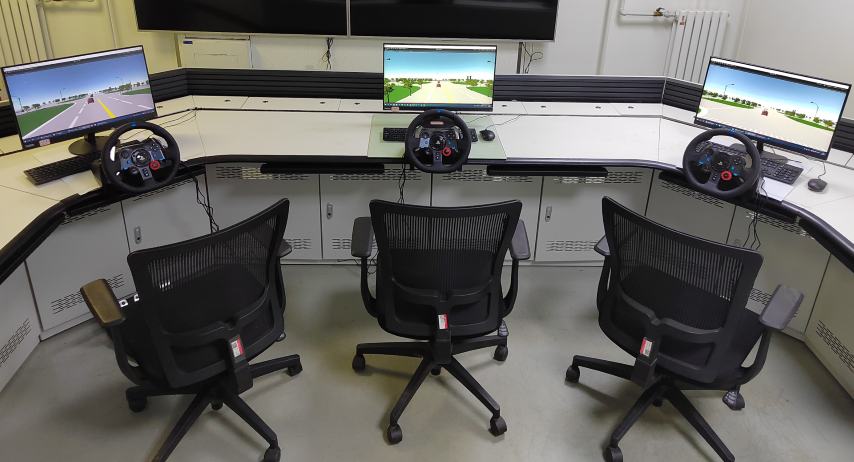}
		\label{fig.g29 simulators deployment}}
	\subfigure[InnoSimulation simulator]
	{\includegraphics[scale=0.3]{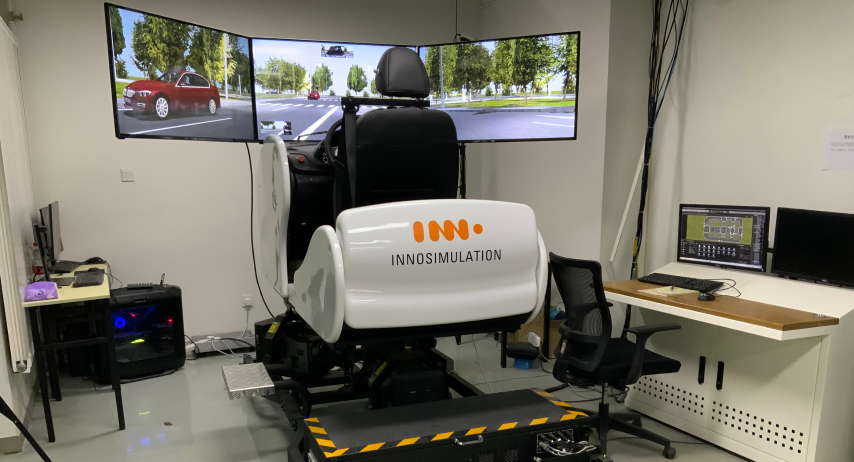}
		\label{fig.InnoSimulation}}
	\vspace{-1mm}
	\caption{Driving simulators with various-fidelity levels. (a) The three Logitech G29 Driving Force simulators, each of which comprises one steering wheel, an accelerator pedal, a brake pedal and a host computer. (b) The high-fidelity commercial driving simulator from InnoSimulation, consisting of a driver cabin, three screens, a motion platform and a host computer.}
	 \vspace{-4mm}
	\label{fig.moniqi}
\end{figure}

\subsection {Virtual Platform}

The virtual platform developed by the Unity game engine represents the virtual space in mixedDT and primarily consists of virtual vehicles with simulation dynamics and virtual roadside facilities.
The virtual driving environment reproduces the physical driving environment in Unity with a $14:1$ scaling ratio, thus having a consistent size as real-world roads; see Fig.~\ref{fig.virtual environment} for demonstration. The update frequency of the entire scene in the virtual platform is $50\,\mathrm{FPS}$.

The virtual vehicles exist independently in the virtual platform without physical counterparts, while the classical DT commonly reproduces the physical vehicles in a digital platform (see, \emph{e.g.},~\cite{chalaki2022research,goldfain2019autorally}).
Due to the independent existence of the virtual vehicles, their quantity, motion behavior and dynamics can be arbitrarily designed. 
Still, they can naturally be designed to serve as digital replicas of the physical vehicles, corresponding to the twin vehicles in the classical DT. 
The typical bicycle model~\cite{rajamani2011vehicle} is utilized as the default dynamics model; see the right panel of Fig.~\ref{fig.virtual vehicle} for demonstration. 
Two commonly used vehicle body models are also presented in Fig.~\ref{fig.virtual vehicle}.
In addition, unlike the physical vehicles, obtaining real-time and accurate states of the virtual vehicles in the virtual platform is highly convenient.

\begin{figure*}[t!]
	\centering
	\includegraphics[scale=0.72]{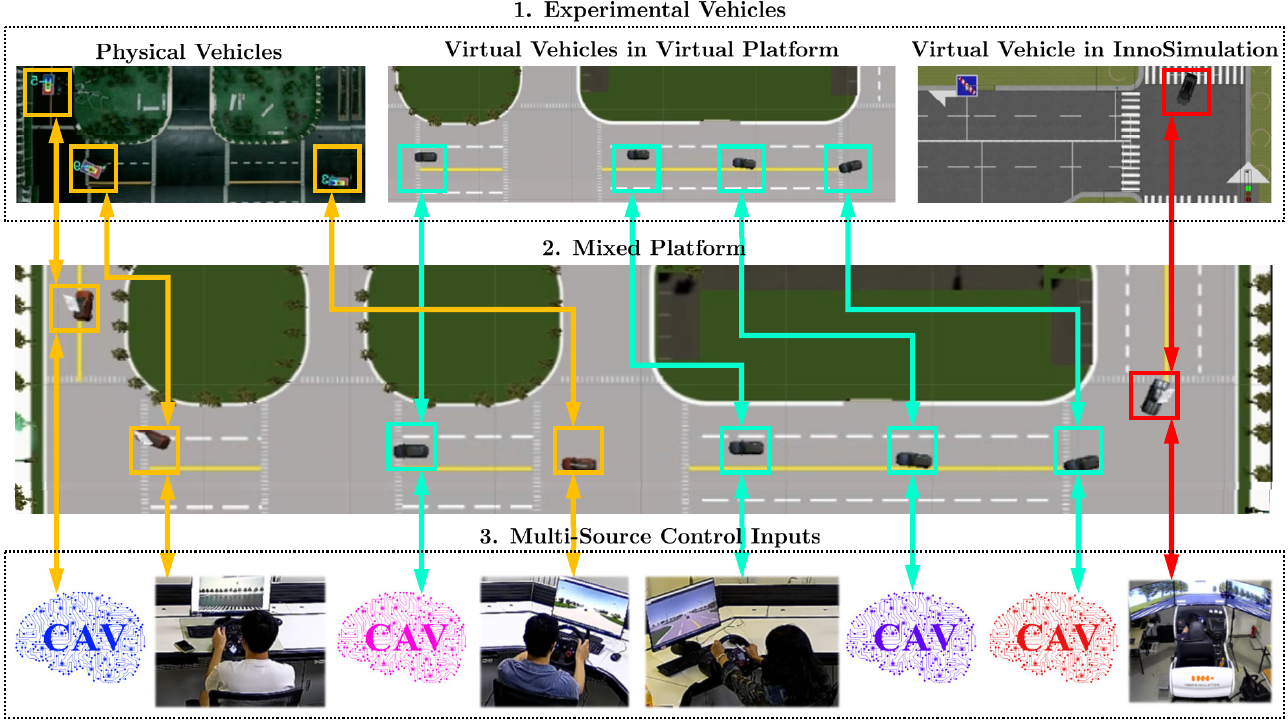}
	\caption{A demonstration diagram of practical MSH-MCCT operation. Multiple vehicles in diverse platforms and environments, controlled by multiple human drivers and CAV algorithms, interact with each other simultaneously via the mixed platform. The visualization of the mixed platform projected to virtual space by Unity is presented on the middle panel. Precisely, there are three miniature physical vehicles from the physical platform (outlined in gold), four virtual vehicles from the virtual platform (outlined in cyan), and one virtual vehicle from the InnoSimulation environment (outlined in red). For the multi-source control inputs, three human drivers on the G29 simulators and one human driver on the InnoSimulation simulator collectively control the four vehicles, while the other four vehicles are controlled by CAV algorithms. The detailed driving mode setups are illustrated in Table~\ref{tab.hdv_setup}.
    }
	\label{fig.mixed platform}
	 \vspace{-4mm}
\end{figure*}

\subsection{Multi-source Control Inputs}
The above presents the controllable physical and virtual vehicles in MSH-MCCT. In the following, we introduces the details of the generation and transmission of control instructions, categorized by their sources into human drivers and CAV algorithms. 

\subsubsection{Huamn Drivers}

Human driver's driving actions are measured and transmitted by the driving simulator, including the steering wheel angle, the accelerator pedal signal and the brake pedal signal. 
In MSH-MCCT, two types of driving simulators are deployed: three Logitech G29 Driving Force simulators, and one high-fidelity commercial driving simulator from InnoSimulation; see Fig.~\ref{fig.moniqi} for illustration. 
In theory, our methodology can support the integration of an arbitrary number of driving simulators with various-fidelity levels in one DT testing environment.

The G29 simulator depicted in Fig.~\ref{fig.g29 simulators deployment} comprises one steering wheel with turning force feedback, an accelerator pedal, a brake pedal and a host computer. 
The steering wheel has a rotation range of $900$ degrees, identical to the real cars. The sampling resolution of the steering wheel and the two pedals are all $65535$. Their values are retrieved through the official SDK from Logitech. 

As shown in Fig.~\ref{fig.InnoSimulation}, the InnoSimulation simulator consists of a driver cabin, three screens, a motion platform and a host computer. 
The driver cabin fully replicates the configuration of a real vehicle's driver seat, including the steering wheel, accelerator and brake pedals, gear shifter, dashboards, seat belt, and handbrake, etc., providing the driver an close feeling to driving a real vehicle. 
The three screens provide drivers an immersive driving field-of-view. Particularly, the scenes from the rear-view mirrors are displayed on three specific regions of the middle screen, corresponding to the left, right, and upper rear-view mirrors of a real car. 
The motion platform is driven by motors and located beneath the driver cabin, providing haptic feedback during driving, and thus enabling a more realistic driving experience. 

Human-in-the-loop experiments via the InnoSimulation simulator are supported by a dedicated commercial software, \emph{i.e.}, SCANeR Studio, which accomplishes the measurement and transmission of driver input signals. 
Moreover, SCANeR Studio is an independent and comprehensive simulation software with extensive Application Programming Interfaces (APIs) similar to Unity. To fully leverage the InnoSimulation simulator hardware via the APIs and enrich driving environment in MSH-MCCT, we construct another virtual driving environment in SCANeR Studio replicating the physical sand table at a $14:1$ scale. This unique driving environment is represented as ``InnoSimulation environment'' in the following for simplicity. 
When operating the InnoSimulation simulator, in addition to driving the vehicles in the physical and virtual platforms similar to the G29 simulators, the driver can also operate the virtual vehicles in the InnoSimulation environment, 
which can be integrated into the mixed platform in a similar manner to the virtual vehicles in the virtual platform. 


\subsubsection{CAV Algorithms}
The CAV algorithm is the main testing subject in MSH-MCCT, and both single-vehicle and multi-vehicle algorithms are supported. The experimenter has the option to run the algorithms directly on the cloud or on any external computer capable of communicating with the cloud, such as their personal computers. In the latter case, to conduct experiments, the external computer only needs to communicate with the cloud according to a pre-defined protocol to obtain the states of all vehicles, and then return control instructions in a requested format. Moreover, since the cloud communication module mainly utilizes the open source cross-language network library ZeroMQ, algorithms developed in almost all mainstream programming languages are supported to access MSH-MCCT, such as C, C++, Java and Python. 

This high level of compatibility enables researchers to conduct algorithm validation experiments quickly and conveniently. 
Human control instructions are transmitted separately by multiple driving simulators to the cloud for subsequent processing. Similarly, CAV algorithms can also run distributedly in multiple computers. 
Besides, since an algorithm can simultaneously control any number of vehicles, multiple CAV algorithms can flexibly match the number of vehicles they control. 
In addition, the lateral and longitudinal control of each vehicle are also decoupled, which supports flexible testing of a dedicated algorithm. 
For example, experimenters can only control the longitudinal movement of one vehicle using their own algorithms, while the lateral movement of this vehicle and other CAVs are controlled by algorithms deployed on the cloud or external computers. 

\subsection {Mixed Platform}
\label{sec.mixed platform}
The mixed platform corresponds to the mixed space in mixedDT and is directly operating on the cloud server. 
It is responsible for system-wise data fusion, control and scheduling, and provides a unified access for external devices to MSH-MCCT. 
Recall that the mixed space in mixedDT is an integration of the physical and virtual spaces, where physical and virtual entities coexist and interact in real-time. 
In MSH-MCCT, the physical platform in the physical space, the virtual platform in the virtual space, and the multi-source control inputs in the physical and virtual spaces are integrated into the mixed platform in the mixed space, through which they indirectly interact with each other. 

We present one example in Fig.~\ref{fig.mixed platform}. 
Three physical vehicles in the physical platform, four virtual vehicles in the virtual platform and one virtual vehicle in the InnoSimulation environment are all running in one unified environment, \emph{i.e.}, the mixed platform; see the upper and middle panels of Fig.~\ref{fig.mixed platform}. 
The vehicles' state data are transmitted to the multi-source control inputs through the cloud. 
The control inputs consist of multi-source human drivers and diverse CAV algorithms; see the lower panel of Fig.~\ref{fig.mixed platform}. 
These control inputs are transmitted by the cloud to the corresponding vehicles for execution and then generate ``control output''. 
In this way, MSH-MCCT operates as one integrated and comprehensive system. 

Specifically, the cloud obtains the state information of the physical vehicles after fusing the measured data from roadside and on-board sensors, while the state information of the virtual vehicles is naturally available to the cloud. 
Due to the unified design among the physical, virtual, and  InnoSimulation environments, the cloud can quickly and accurately align the state data of all types of vehicles. 
Then, the cloud projects the location and motion states of all the vehicles into a unified environment, \emph{i.e.}, the mixed platform. 
As a result, the state data from all the vehicles are aggregated in the cloud into a standardized format data pool, accessible to any authorized entity, such as the multi-source control inputs. 

Meanwhile, the cloud receives, transforms, and sends the multi-source control inputs of human drivers and CAV algorithms to the corresponding vehicles for execution, with an update frequency of $50\,\mathrm{Hz}$. Since the control instructions can be directly sent to the corresponding vehicles for execution, they do not need to be aggregated together like the vehicle state data. Therefore, the cloud processes the control instructions separately according to their corresponding execution vehicles, which makes full use of the abundant multi-thread computational resources in the cloud to improve processing speed. 
The control instructions are first converted to the scale and type of the corresponding vehicle, and then transformed into specific formats after appropriate filtering. Subsequently, they are sent to the corresponding vehicles for execution according to specific protocols. 
Specifically, the instructions to the physical vehicles are converted into ROS message format and then sent to different physical vehicles via ROS. The instructions to the virtual vehicles in the virtual platform are converted into a predefined JSON format and then sent to the virtual platform via ZeroMQ. Then, the status of all types of the vehicles are updated accordingly. 

Consequently, the behaviors of physical and virtual HDVs can truly affect those of physical and virtual CAVs, and vice versa. 
In other words, four types of vehicles from diverse platforms and environments coexist and interact with each other indirectly via the mixed platform.

\begin{remark}[Integration of diverse platforms and environments]
Based on mixedDT, MSH-MCCT overcomes the typical ``organizational siloing'' problem for DT applications~\cite{grieves2017digital},  which refers to the isolated existence and operation of multiple organizations within the system, lacking information sharing, interaction and integration. Precisely, it bridges the vehicles in the physical and virtual platforms, and the multi-source control inputs from multiple driving simulators and CAV algorithm hosts to operate together simultaneously within one unified and integrated environment, \emph{i.e.}, the mixed platform. 
The boundaries between these originally independent and separate platforms and environments are eliminated in MSH-MCCT, and synchronous operation and cross-platform real-time interaction are achieved; see Fig.~\ref{fig.mixed platform} for illustration. 
In this manner, one can fully explore and utilize the advantages of the diverse platforms and environments based on mixedDT and thus realize more potential applications than existing platforms~\cite{wang2021digital,la2012development,kannapiran2020go}. 

\end{remark}

\subsection{Visualization, Communication, and Interaction}
\label{sec.visualization, communication and hmi}

Since the mixed space is an integration of the physical and virtual spaces, obtaining a comprehensive visualization of the mixed platform is not straightforward. 
In MSH-MCCT, we design two approaches: 1) First is visualization projected to the physical space by a commercial MR head-mounted display, the HoloLens. Fig.~\ref{fig.MSH-framework}(c) presents the image obtained by directly photographing one lens of the HoloLens in the physical space. 
2) Second is visualization projected to the virtual space. Since the cloud contains state data of all vehicles in standardized format, any virtual environment authorized to access the cloud can realize such visualization, such as Unity, SCANeR Studio, SUMO, and CARLA. The middle panel of Fig.~\ref{fig.mixed platform} exemplifies the visualization by Unity. 

As shown in~Fig.~\ref{fig.control mode}, the communication between the cloud and other components in MSH-MCCT and external devices mainly adopts the open source cross-language network library ZeroMQ with TCP protocol, through which the cloud can establish interaction with various entities. 
Specifically, wireless connection mode is $5\mathrm{GHz}$ WiFi, and wired connection mode is Ethernet. 
Note that the communication between the driving simulators and the cloud, corresponding to link \circled{1} in~Fig.~\ref{fig.control mode}, is established by reliable and high-speed wired connection, with an average latency of $0.66\; \mathrm{ms}$. The communication latencies of the other links are reported in~\cite{dong2023mixed}.

Additionally, many HMI applications can be implemented via MR devices, such as the HoloLens, which allow users to interact with the entities in MSH-MCCT. 
For example, we can put an obstacle at a specific position at anytime to test obstacle avoidance algorithms, and change the velocity of a specific vehicle to impose external perturbations during experiments. Interested readers are referred to~\cite[Section IV]{dong2023mixed} for details.

\begin{figure}[t]
	\vspace{1mm}
	\centering
	\includegraphics[scale=0.36]{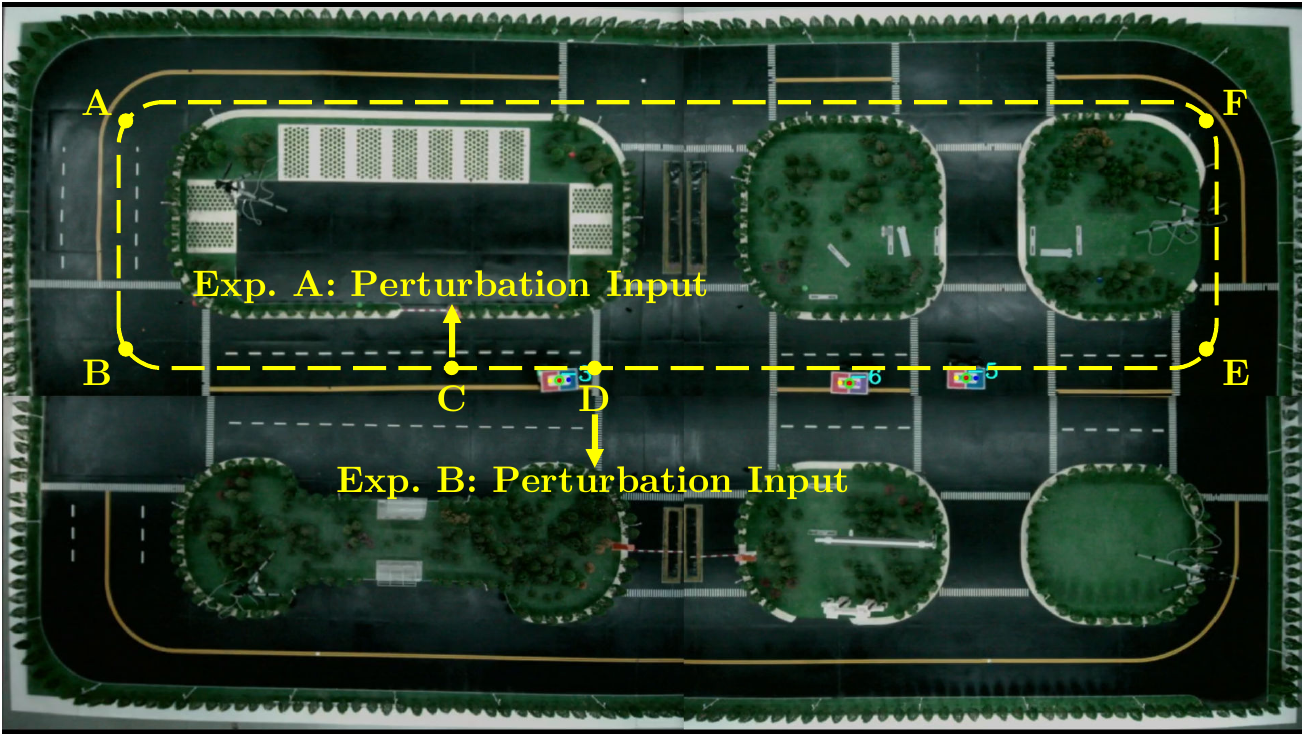}
	\caption{The vehicles drive along the direction $A \rightarrow B\rightarrow C\rightarrow D\rightarrow E\rightarrow F\rightarrow A$. In Experiment~A (traffic wave scenario), when reaching point $C$, the velocity of the head vehicle suffers from a half-sinusoidal perturbation. In Experiment~B (safety-critical scenario), when reaching point $D$, a braking perturbation is imposed on the head vehicle. 
    }
	\label{fig.track}
  \vspace{-4mm}
\end{figure}

\begin{table*}[t]
\footnotesize
\caption{Driving Mode Setups}
\label{tab.hdv_setup}
\centering
		\begin{threeparttable}
		\setlength{\tabcolsep}{4mm}{
			\begin{tabular}{ccccc}
   \toprule
    ID & Type & Entity & Control input & Field-of-View   \\\hline
    1 & CAV & Physical vehicle & CACC controller & / \\
    2 & HDV & Physical vehicle & Human via G29 simulator & Physical view from the sand table \\
    3 & CAV & Virtual vehicle & CACC controller & / \\
    4 & HDV & Physical vehicle & Human via G29 simulator & Virtual view from the Unity environment\\
    5 & HDV & Virtual vehicle & Human via G29 simulator & Virtual view from the Unity environment \\
    6 & CAV & Virtual vehicle & CACC controller & / \\
    7 & CAV & Virtual vehicle & CACC controller & / \\
    8 & HDV & Virtual vehicle & Human via InnoSimulation simulator & Virtual view from the InnoSimulation environment \\
				\bottomrule
		\end{tabular}}
		\end{threeparttable}
\end{table*}

\begin{figure*}[t]
	\vspace{1mm}
	\centering
	\subfigure[Physical platform captured by roadside cameras]
	{\hspace{0.1mm}
 \includegraphics[scale=0.35]{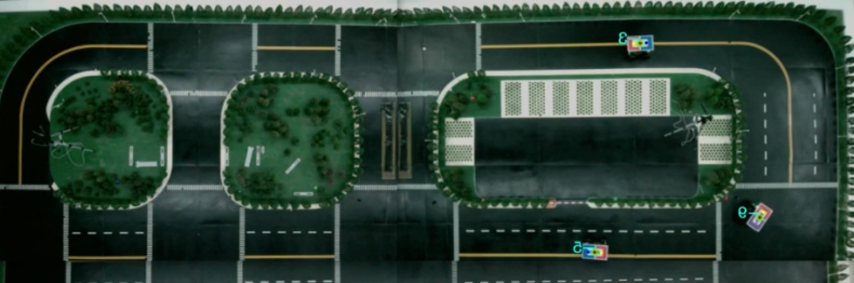}
		\label{fig.roadside cameras}}
	\subfigure[Visualization of the mixed platoon by Unity]
	{\includegraphics[scale=0.35]{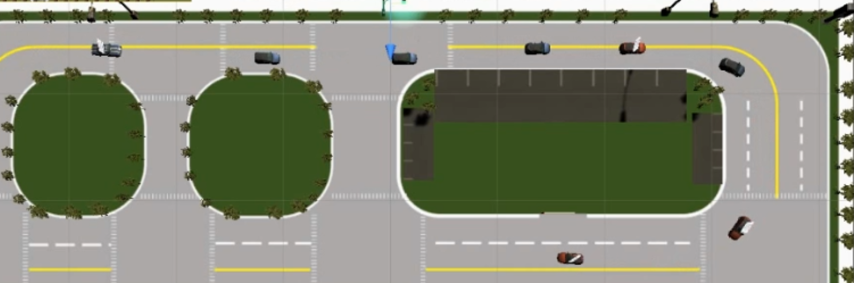}
		\label{fig.vis-mixed}}
    \subfigure[Three G29 simulators]
    {    \hspace{0.1mm}
    \includegraphics[scale=0.35]{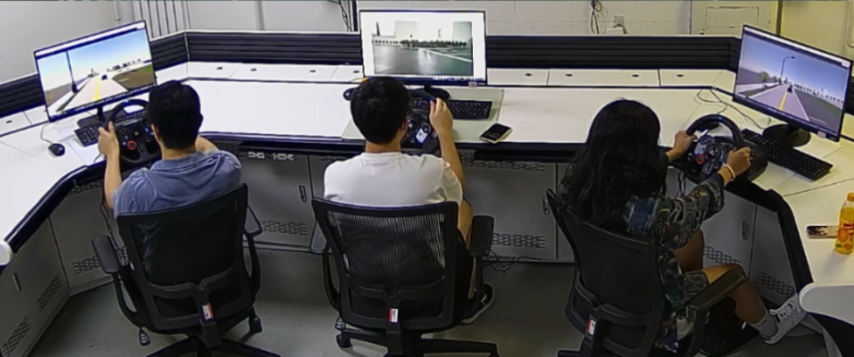}
		\label{fig.G29simulators}}
  \subfigure[InnoSimulation simulator]
	{\includegraphics[scale=0.35]{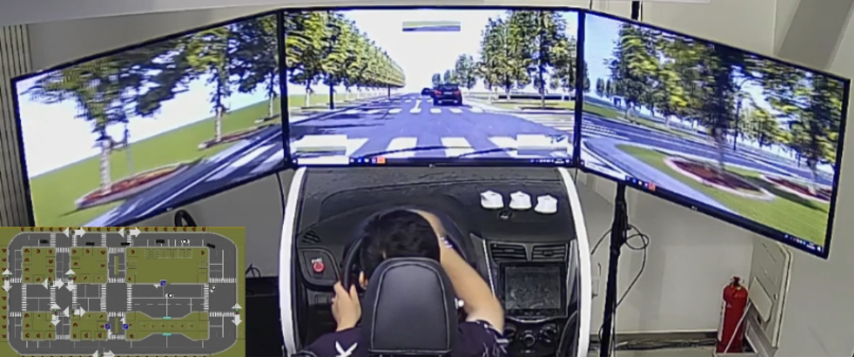}
		\label{fig.INNO simulators}}
	\vspace{-1mm}
	\caption{Snapshots of the platform effectiveness experiment. In (a), the image captured by the roadside cameras in the physical platform displays running physical vehicles. In (b), the visualization of the mixed platoon by Unity is presented, with the coexistence of four HDVs and four CAVs. The corresponding driving mode setups of the platoon are shown in Table~\ref{tab.hdv_setup}. In (c) and (d), three male and one female drivers with various levels of driving experiences operate vehicles via different-fidelity driving simulators. Particularly, as can be observed from the screens in (c), the driver in the middle is driving based on physical field-of-view, while the other two are driving based on virtual field-of-view. In (d), the driver is driving based on virtual field-of-view from the InnoSimulation environment, whose overview is shown in the lower left corner of (d). The corresponding videos can be found in \url{https://dongjh20.github.io/MSH-MCCT}.
	}
	\label{fig.snapshot}
  \vspace{-4mm}
\end{figure*}

\section{Experimental Validations}
\label{sec.5}
In this section, we conduct experiments on vehicle platooning in mixed traffic, \emph{i.e.}, mixed platooning with the coexistence of CAVs and HDVs, to validate the effectiveness of MSH-MCCT. Existing research on mixed platooning mostly utilize traffic simulations for validation, where the HDVs' dynamics are captured by typical car-following models; see, \emph{e.g.},~\cite{wu2021flow,zheng2020smoothing}. In the following, we aim to demonstrate the capability of MSH-MCCT for conducting CAV testing with multi-source real human drivers in the loop.  
Particularly, a traffic wave scenario and a safety-critical scenario are under consideration. Note that in this section, all the units have been aligned with the real-world environment. 
As demonstrated in~\cite{wang2023implementation}, MSH-MCCT can reproduce the phenomenon of stop-and-go traffic waves~\cite{treiber2013traffic} in fully HDV conditions not only in the straight-road scenario under external disturbances but also in the ring-road scenario with no bottlenecks, demonstrating its capability to replicate real-world traffic environments. 
For more MSH-MCCT-based research, see~\cite{li2025mixed,li2025robust}.

\begin{figure*}[t]
	\centering
	\subfigure[Experiment~A: traffic wave scenario]
	{\includegraphics[scale=0.35]{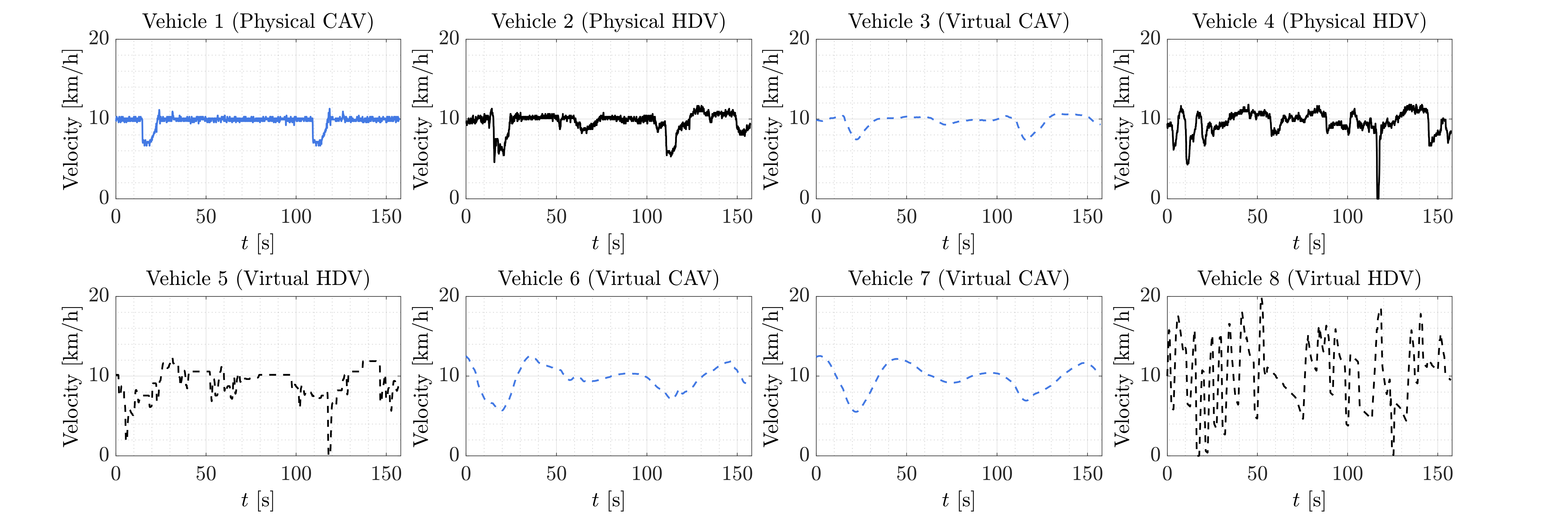}
		\label{fig.speed-A}}
	\subfigure[Experiment~B: safety-critical scenario]
	{\includegraphics[scale=0.35]{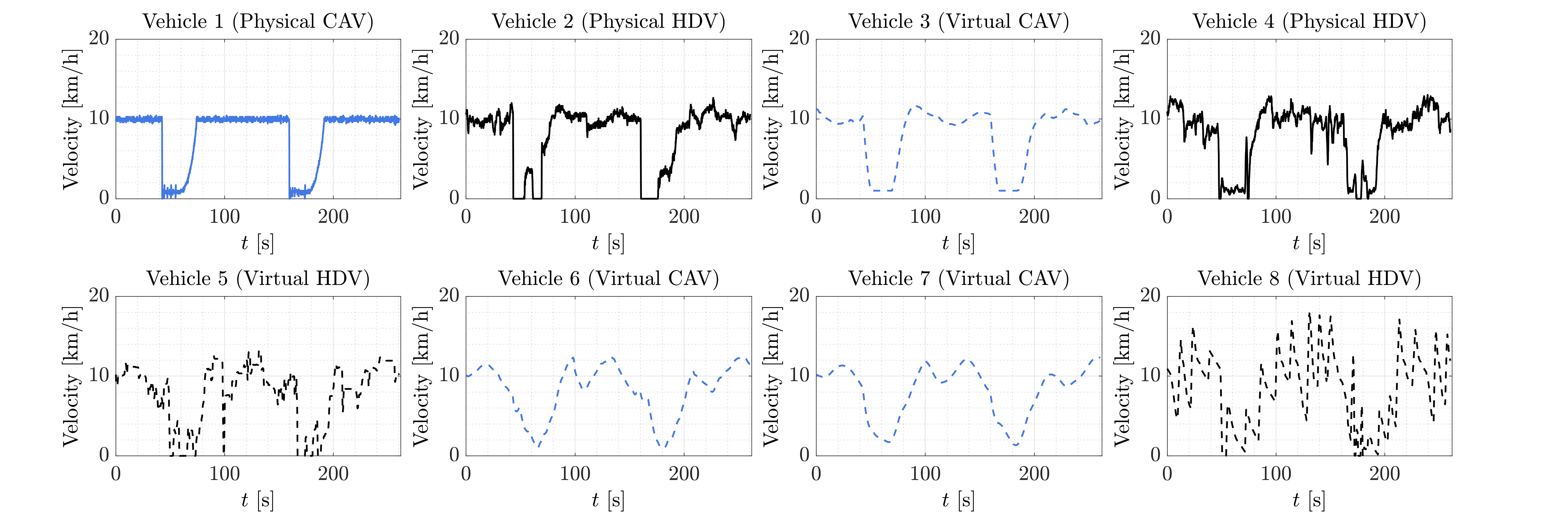}
		\label{fig.speed-B}}
	\vspace{-1mm}
	\caption{Velocity profiles of all the vehicles in the platform effectiveness experiments. The physical vehicle is in solid line, while the virtual vehicle is in dashed line. The CAV is in blue line, while the HDV is in black line. In Experiment~A, a half-sinusoidal perturbation is imposed to the head vehicle, while in Experiment~B, a braking perturbation is imposed. The velocity values have been unified to the virtual platform, which has a consistent size as real-world roads. 
	}
	\label{fig.speed}
 \vspace{-4mm}
\end{figure*}

\begin{figure*}[t]
	\centering
	\subfigure[Experiment~A: traffic wave scenario]
	{\includegraphics[scale=0.35]{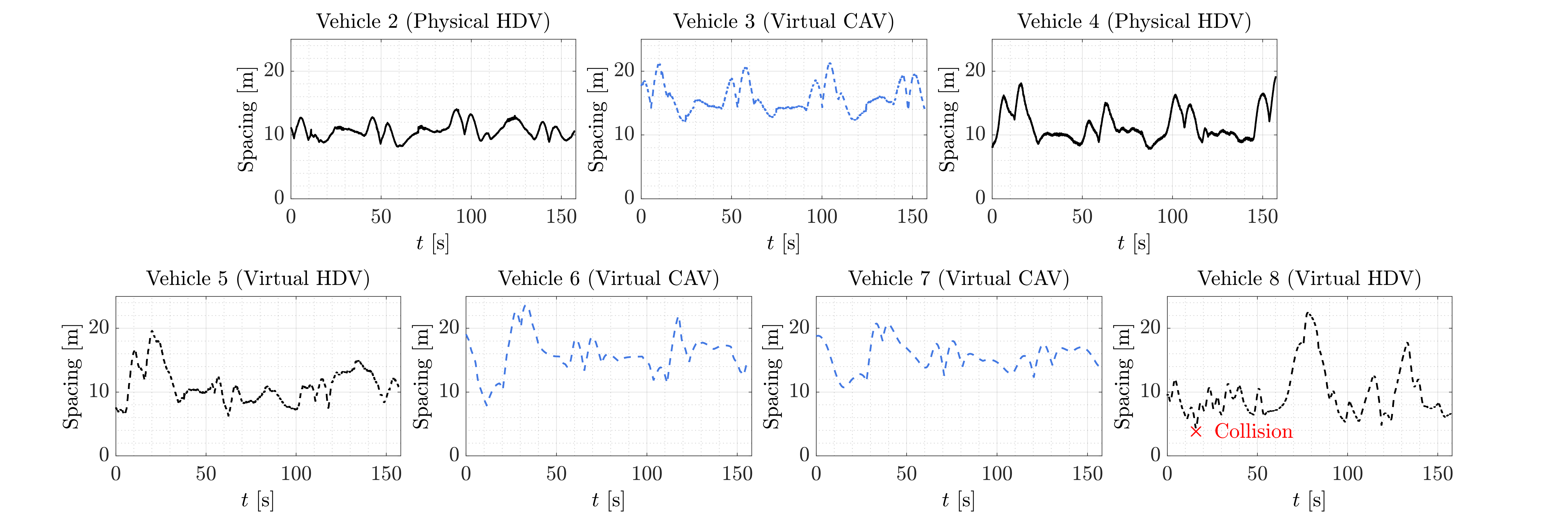}
		\label{fig.distance-A}}
	\subfigure[Experiment~B: safety-critical scenario]
	{\includegraphics[scale=0.35]{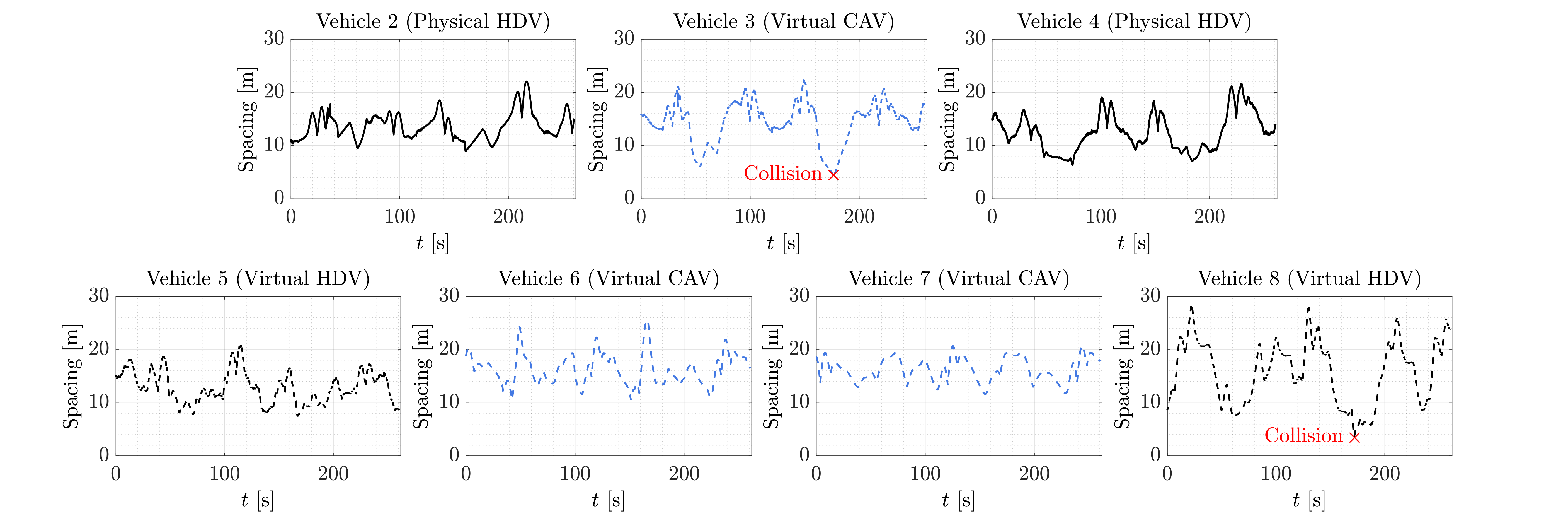}
		\label{fig.distance-B}}
	\vspace{-1mm}
	\caption{Spacing profiles of all the following $7$ vehicles in the platform effectiveness experiments. The inter-vehicle spacing here represents the distance between the centroids of the two vehicles. The physical vehicle is in solid line, while the virtual vehicle is in dashed line. The CAV is in blue line, while the HDV is in black line. Particularly, three collisions indeed occur, marked with red crosses. However, since at least one of the involved participants is a virtual vehicle, no actual damage is incurred by any vehicle. This shows that our platform supports safety-critical scenarios when testing CAV technologies. 
	}
	\label{fig.distance}
  \vspace{-4mm}
\end{figure*}

\subsection{Experimental Setup}

{In our experiments, a platoon of $8$ vehicles, consisting of one head vehicle and $7$ following vehicles, are deployed to run along the track $A \rightarrow B\rightarrow C\rightarrow D\rightarrow E\rightarrow F\rightarrow A$ in Fig.~\ref{fig.track} with a lap of about $245\,\mathrm{m}$. 
Two scenarios are designed: 1) Experiment~A:  traffic wave scenario. The head vehicle maintains a fixed velocity of $10.08\,\mathrm{km/h}$ during the driving track, and a half-sinusoidal perturbation occurs when it reaches point $C$ shown in Fig.~\ref{fig.track}. The perturbation period is $3.50\,\mathrm{s}$, and its amplitude is $3.02\,\mathrm{km/h}$. 2)  Experiment~B: safety-critical scenario. The head vehicle still maintains a fixed velocity of $10.08\,\mathrm{km/h}$, but when it reaches point $D$, a braking perturbation is imposed, which consists of three phases: decelerate to $1.01\,\mathrm{km/h}$ by a rate of $0.28\,\mathrm{m/s^2}$, maintain this speed for $20\,\mathrm{s}$, and finally return to $10.08\,\mathrm{km/h}$ after $12\,\mathrm{s}$. In both experiments, the platoon operates steadily for four cycles. Additionally, the two experiments are conducted with a 15-minute interval to provide human drivers with sufficient rest time.}

The detailed driving mode setups for the platoon are illustrated in Table~\ref{tab.hdv_setup}, where there are three physical vehicles and five virtual vehicles, among which one is from the InnoSimulation environment, and the other four are from the virtual platform. Note that in the experiments, the formation pattern is also the same as that of the example shown in Fig.~\ref{fig.mixed platform}. Precisely, the HDVs and the CAVs are controlled as follows:

\begin{itemize}
    \item 
For the HDVs, the multi-source human drivers consist of three male drivers with different driving experiences (five, ten and twenty years respectively), and one female driver with five-year driving experience. 
Before the formal experiments, they all familiarize themselves with operating the vehicle via the specific driving simulator, and are informed about the details of the two experimental scenarios. 
Besides, they are advised to adopt a relatively conservative driving style to avoid potential collisions. The detailed driving mode setups for the four human drivers are illustrated within Table~\ref{tab.hdv_setup}.

\item 
For the CAVs, the lateral and longitudinal control are decoupled, and the control algorithms are deployed directly on the cloud. 
Precisely, we utilize a typical preview trajectory tracking controller~\cite{amer2017modelling} for the lateral control. For the longitudinal control, the following CACC controller is adopted~\cite{milanes2013cooperative}
\begin{equation}
	\label{equ.cacc}
	\begin{aligned}
		a_i(t)=&k_{p}\left((p_i(t)-p_{i-1}(t))-d_\mathrm{des}\right)+k_{v1}(v_{1}(t)-v_i(t)) \\ &+k_{v2}(v_{i-1}(t)-v_i(t)),
\end{aligned}
\end{equation}
where the control signal $a_i$ represents the desired acceleration of vehicle $i$, and $k_{p}$, $k_{v1}$ and $k_{v2}$ are all feedback gains. The state variables $p_{i}$ and $v_{i}$ denote the longitudinal position and velocity of vehicle $i$ respectively, and $d_\mathrm{des}$ represents the desired inter-vehicle spacing. 
Since the standard longitudinal control instruction is the desired velocity, the desired acceleration $a_i(t)$ of each vehicle is converted to the velocity command signal $v_{i,\mathrm{cmd}}(t)$ in the cloud by
\begin{equation*} \label{Eq:CommandVelocity}
    v_{i,\mathrm{cmd}}(t) = v_{i}(t-1) + a_i(t) \Delta t,
\end{equation*}
where $v_{i}(t-1)$ represents the received velocity of vehicle $i$ at the last time step $t-1$, and $\Delta t$ represents the time interval between $t$ and $t-1$. 
\end{itemize}

\subsection{Experimental Results}

We proceed to show the experimental results. The snapshots of the two experiments are shown in Fig.~\ref{fig.snapshot}.  From the experiments, the velocity and spacing profiles of each vehicle are shown in Fig.~\ref{fig.speed} and Fig.~\ref{fig.distance}, respectively.
It can be clearly observed that the platoon, composed of vehicles from diverse platforms and environments, run normally in the mixed platform. 
In Experiment~A (traffic wave scenario), the CACC controllers enable the CAVs to dampen the velocity fluctuation of the head vehicle, preventing disturbances from amplifying within the platoon, while the human drivers are amplifying these fluctuations, especially the last HDV. 
In Experiment~B (safety-critical scenario), due to the sudden brake of the head vehicle, a larger speed fluctuation is introduced into the platoon, leading to more significant speed fluctuations within the entire platoon. 
Still, the CAVs distributed in the platoon mitigate the amplification of fluctuations, and the speed of the entire platoon does not diverge. Our multi-source human-in-the-loop experiments validate the practical wave-dampening capability of CACC, which is consist with the findings from the real-world test~\cite{milanes2013cooperative}. 

Moreover, although the drivers are instructed to adopt a conservative driving style, three collisions still occur due to the head vehicle's sudden speed fluctuations; see Fig.~\ref{fig.distance} for illustration. The minimum distance between the centroids of the two vehicles are $4.23\,\mathrm{m}$, $4.52\,\mathrm{m}$ and $3.67\,\mathrm{m}$, respectively, marked with red crosses in Fig.~\ref{fig.distance}.
Considering the actual physical dimensions of the vehicle body, three collisions have indeed occurred. This result reveals the limited safety performance of CACC in practice with respect to the entire mixed traffic flow, where rear-end collisions could still happen somewhere in traffic with $50\%$ CAVs. This safety-related result is non-trivial to be reproduced in real traffic given the real-world cost, and is poor in fidelity in simulations with no real human drivers in the loop. 
Note that in our experiments all the collisions involve at least one virtual vehicle, and thus no actual damage is incurred by any vehicle. This  demonstrates that our platform supports safety-critical scenarios when testing CAV technologies. 

In addition, we also observe from Fig.~\ref{fig.speed} and Fig.~\ref{fig.distance} that the velocity profile of the physical vehicle has apparently larger oscillations  than that of the virtual vehicle. This is due to the unknown noise and disturbance in the real-world physical environment, which cannot be completely replicated in the digital simulators. 
This is consistent with the significance of establishing our mixed platform: to fully explore and utilize the advantages of diverse platforms and environments. 
These experimental results validate the effectiveness of MSH-MCCT for conducting CAV testing with multi-source real human drivers in the loop, and the capability of MSH-MCCT for supporting real-time cross-platform interaction.

\section{Conclusion}
\label{sec.6}
In this paper, we introduce our experimental platform MSH-MCCT, which is established based on the notion of mixedDT and serves for conducting CAV testing with multi-source real human drivers in the loop via various-fidelity driving simulators. The four components of MSH-MCCT (physical, virtual, and mixed platforms, and multi-source control inputs) achieve synchronous operation and cross-platform real-time interaction. Consequently, physical and virtual CAVs and HDVs could coexist and interact within an integrated environment, greatly enhancing the experimental flexibility and scalability. Particularly, we organize multiple vehicles from diverse platforms and environments into one platoon in mixed traffic, and further conduct experiments in the traffic wave and safety-critical scenarios. 
The results validate the capability of MSH-MCCT for CAV testing while integrating multi-source real human drivers in the loop. 

Based on MSH-MCCT, multiple topics deserve future investigation. Beyond platooning, various CAV applications, such as multi-lane formation~\cite{cai2022formation}, intersection control~\cite{wang2023connected}, and on/off-ramps cooperation~\cite{liao2021cooperative}, can be tested with interaction between multiple CAVs and HDVs. Considering that roadside facilities such as traffic lights also play a crucial role in transportation system, another future direction is to explore vehicle-road cooperation involving multiple CAVs, HDVs and intelligent roadside facilities. Besides CAV testing, the behavior pattern of human drivers in the case of CAV existence can also be captured and investigated via MSH-MCCT by collecting various physiological signals~\cite{wu2021inferring}, which is believed to differ from human's normal driving behaviors without CAVs~\cite{soni2022behavioral}. 
Finally, the cloud constitutes the core of MSH-MCCT, and how to construct a more powerful and flexible mixed platform based on the multi-layered cloud architecture is also an interesting topic.


\ifCLASSOPTIONcaptionsoff
  \newpage
\fi



%

\bibliographystyle{IEEEtran}
\bibliography{IEEEabrv,mybibfile}

\end{document}